\documentclass[10pt, a4paper]{article}

\usepackage{hyperref}
\usepackage{times}
\usepackage{latexsym}
\usepackage{multirow}
\usepackage{booktabs}
\usepackage{graphicx}
\usepackage{amssymb}
\usepackage{makecell}
\usepackage{amsmath, bm}
\usepackage{longtable}
\usepackage{subfigure}
\usepackage{caption}
\usepackage{enumerate}
\usepackage{enumitem}
\usepackage{algorithm}
\usepackage{algorithmic}
\usepackage{longtable}
\usepackage{xspace}
\usepackage{comment}
\usepackage{color}
\usepackage{colortbl} 

\usepackage{bbding}

\usepackage{pifont}       
\usepackage{bbding}       
\usepackage{fontawesome}  

\usepackage[tikz]{bclogo}

\definecolor{green1}{RGB}{215,228,190}
\definecolor{red1}{RGB}{242,217,217}
\definecolor{orange1}{RGB}{255,241,221}
\definecolor{blue1}{RGB}{222,240,250}

\usepackage[]{lrec-coling2024} 

\title{\textbf{Prompting Large Language Models for Counterfactual Generation: An Empirical Study}}

\name{Yongqi Li$^{1,*}$, Mayi Xu$^{1,*}$, Xin Miao$^{1,*}$, Shen Zhou$^{1,*}$, Tieyun Qian$^{1,2,\dagger}$}
\address{$^1$ School of Computer Science, Wuhan University, China \\
        $^2$ Intellectual Computing Laboratory for Cultural Heritage, Wuhan University, China\\
         \{liyongqi, xumayi, miaoxin, shenzhou, qty\}@whu.edu.cn\\}

\abstract{
Large language models (LLMs) have made remarkable progress in a wide range of natural language understanding and generation tasks. 
However, their ability to generate counterfactuals has not been examined systematically. 
To bridge this gap, we present a comprehensive evaluation framework on various types of NLU tasks, which covers all key factors in determining LLMs' capability of generating counterfactuals.
Based on this framework, we 1) investigate the strengths and weaknesses of LLMs as the counterfactual generator, and 2) disclose the factors that affect LLMs when generating counterfactuals, including both the intrinsic properties of LLMs and prompt designing.
The results show that, though LLMs are promising in most cases, they face challenges in complex tasks like RE since they are bounded by task-specific performance, entity constraints, and inherent selection bias.
We also find that alignment techniques, e.g., instruction-tuning and reinforcement learning from human feedback, may potentially enhance the counterfactual generation ability of LLMs. 
On the contrary, simply increasing the parameter size does not yield the desired improvements.
Besides, from the perspective of prompt designing, task guidelines unsurprisingly play an important role.
However, the chain-of-thought approach does not always help due to inconsistency issues.
\\ \newline \Keywords{Large Language Models, Counterfactual Generation, Natural Language Understanding} }

\begin{document}

\maketitleabstract

\renewcommand{\thefootnote}{\fnsymbol{footnote}}
\footnotetext[1]{{ }Equal contributions.}
\footnotetext[2]{{ }Corresponding author.}
\renewcommand{\thefootnote}{\arabic{footnote}}

\section{Introduction}


Counterfactual generation, designed to eliminate spurious correlations in data, is a crucial technique used in causal intervention~\cite{pearl-1993-bayesian-intervention}.
In recent years, many studies~\cite{Kaushik2020Learning,niu2021counterfactual,zhang2023coco} have attempted to enhance the robustness and performance of neural network models through counterfactual generation.

Large language models (LLMs) like ChatGPT are revolutionizing the field of natural language processing~\cite{liu-2023-summary-chatgpt}. 
Due to their power in understanding instructions, learning in context, and text generation, LLMs have attracted widespread attention in utilizing prompt engineering to generate text in specific scenarios.
However, the potential of LLMs in generating counterfactuals remains unexplored systematically. 
This paper aims to bridge this gap by answering two key questions,
1) \textit{What strengths and weaknesses do LLMs have in generating counterfactuals?} 
2) \textit{What factors influence the counterfactual generation ability of LLMs?}



To answer these two questions, we develop a comprehensive framework for evaluating LLMs' capability of generating counterfactuals on four typical natural language understanding (NLU) tasks, i.e., sentiment analysis (SA), natural language inference (NLI), named entity recognition (NER), and relation extraction (RE).
Our framework covers all key factors in LLMs, including the inherent properties of LLMs themselves like the model size as well as the prompt designing for LLMs.  


For the first question, we select the powerful GPT-3.5 as an example for evaluation.
The experimental results show that LLMs can bring about promising enhancements under most settings. 
However, LLMs also have displayed some weaknesses when dealing with complex tasks such as RE.
Further, to discover reasons for the weakness, we first examine the correlation between the quality of generated counterfactuals and the task-specific performance of LLMs.
Then, we explore the factors that are crucial in determining the quality of counterfactuals in the RE task, regarding the satisfaction of entity constraints and the selection bias.

For the second question, we first employ the proposed evaluation framework on Llama-2 family of LLMs~\cite{touvron2023llama2}, which includes {\{7,13,70\}b, \{7,13,70\}b-chat} versions, to investigate the impact of parameter sizes and alignment techniques.
Second, we evaluate GPT-3.5 using different prompt variants to examine whether the task guidelines and chain-of-thought (CoT)~\cite{wei2022chain} are beneficial, and whether the counterfactual generation ability of LLMs is learned from the demonstration or is intrinsic.

Overall, this study makes two major contributions as follows: 

(1) We are the first to present a comprehensive framework for systematically evaluating the counterfactual generation ability of LLMs.
Our framework covers various types of NLU tasks and all key factors in LLMs, including the parameter size, alignment technique, task guideline and CoT.
This framework is then deployed to investigate the strengths and weaknesses of LLMs when generating counterfactuals.

(2) Our study reveals that LLMs can generate high-quality counterfactuals in most cases, but struggle to handle complex tasks such as RE. 
Moreover, the alignment technique can enhance the counterfactual generation capabilities of LLMs, whereas increasing the parameter size or applying CoT is not always beneficial.


\section{Related Work}
\subsection{Large Language Models (LLMs)}

Recently, there have been breakthrough advances in the capabilities of large language models (LLMs)~\cite{zhao-2023-survey-LLMs}, especially in understanding instructions and in-context learning~\cite{2022_icl_survey}. 
The improvement of these capabilities mainly comes from the scaling up of the parameter size, also known as the emergence phenomenon~\cite{wei-2022-LLMs-emergent}, and the inclusion of alignment techniques~\cite{rlhf-2022-openai}, such as instruction-tuning and reinforcement learning with human feedback.
Besides, when prompting LLMs for specific tasks, researchers have also found some ways to improve the performance, such as providing detailed task descriptions~\cite{2020_turking_test}, adopting chain-of-thought~\cite{wei2022chain, 2022_zero_shot_cot} and selecting reasonable demonstration~\cite{liu2022makes}.
In this study, we comprehensively examine these potentially affecting factors of LLMs for counterfactual generation.


\subsection{Counterfactual Generation}

Recent research on causal inference theory~\cite{pearl2009causal,rubin-1974-estimating-causal,morgan-2015-counterfactuals-causal,pearl-2018-book-of-why,feder-2022-causal-nlp} has gained increasing attention due to its potential to enhance the model performance and stability by mitigating spurious correlations in the data~\cite{Kaushik2020Learning,niu2021counterfactual}.
In the area of natural language processing, counterfactual generation has emerged as a prominent area of interest and been employed for various tasks, such as text classification~\cite{garg-ramakrishnan-2020-bae,wang2021robustness}, question answering~\cite{ou-etal-2022-counterfactual-Open-Domain-Dialogues,paranjape-etal-2022-retrieval}, sentiment analysis~\cite{Kaushik2020Learning,ross2021explaining,robeer2021generating,chen2021reinforced,yang2021exploring,howard2022neurocounterfactuals,wen-etal-2022-autocad}, natural language inference~\cite{dixit2022core,wen-etal-2022-autocad}, named entity recognition~\cite{zeng2020cfgen,yang2022factmix}, and relation extraction~\cite{zhang2023coco,miao-2023-generating_commonsense_cf}.
These methods mainly follow the paradigm of causal identification, label-controlled generation and data augmentation, which is also adopted by our proposed evaluation framework.

Notably, there are very limited LLMs-based methods~\cite{dixit2022core,chen2023disco} for counterfactual generation, which only focus on relatively simple SA and NLI tasks.
Moreover, a comprehensive evaluation on LLMs for generating counterfactuals is missing from the literature. 
To fill this gap, we propose an evaluation framework and conduct a multi-perspective empirical study for counterfactual generation using LLMs, covering various types of NLU tasks including SA, NLI, NER and RE.



\section{Methodology}\label{sec:methodology}
\subsection{Causal Theoretical Foundation}\label{sec:causal_formalism}
In this subsection, we use the structural causal model (SCM)~\cite{pearl-2000-models-reasoning-inference} to establish a causal theoretical foundation for counterfactual generation and augmentation.
Here we use the SCM of the SA task as a representative for illustration.

\begin{figure}[h]
\centering
\includegraphics[width=0.48\textwidth]{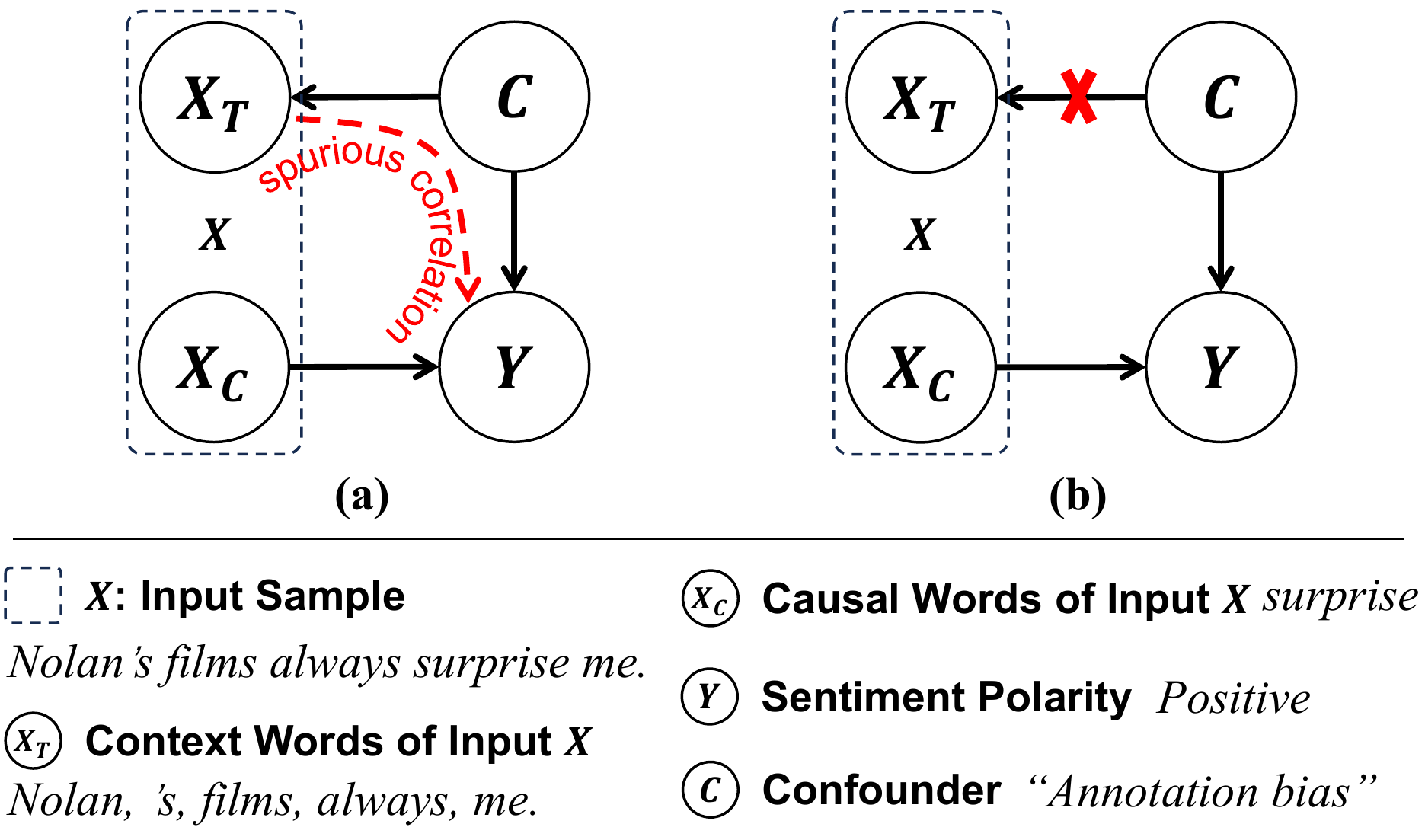}
\caption{(a) Structural causal model of the SA task, (b) Intervention operation.}
\label{fig:causal_formalism}
\end{figure}

\paragraph{Structural Causal Model (SCM)}
As shown in Fig.~\ref{fig:causal_formalism} (a), the SCM mainly shows the relationships among the causal words ($X_C$), the context words ($X_T$) and the sentiment polarity ($Y$).
$X_C\rightarrow Y$: the causal words ($X_C$), i.e., sentiment-related words, are the sole cause of sentiment polarity ($Y$).
$C\rightarrow X_T$ and $C\rightarrow Y$: 
Since only the sentiment-related words $X_C$ are attended during the collection of SA training samples, the distribution of the context words $X_T$ and the sentiment polarity $Y$ is ignored. 
Thus there may be an annotation bias, i.e., the hidden confounder $C$, affecting both $X_T$ and $Y$.

\begin{figure*}[t]
\centering
\includegraphics[width=1\textwidth]{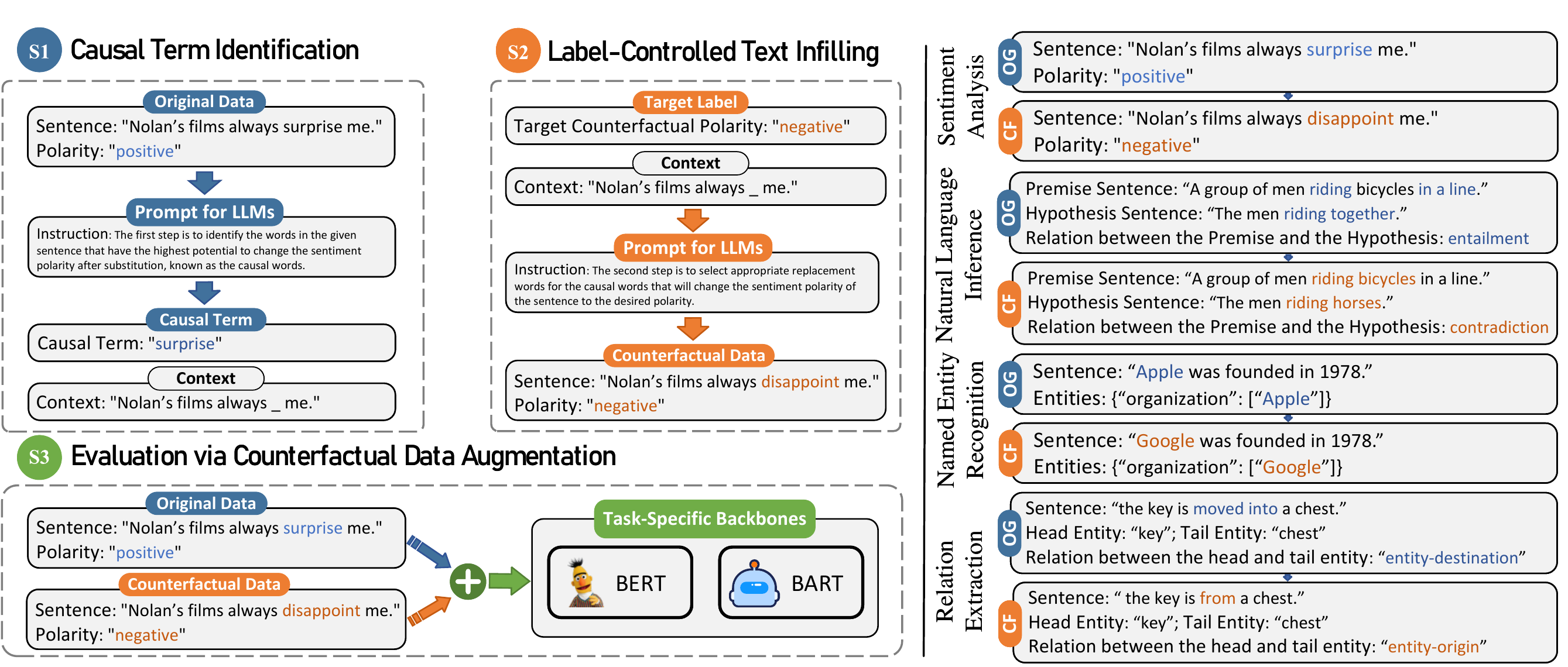}
\caption{Left: The proposed framework for evaluating counterfactuals generated by LLMs (SA task). Right: Original (OG) samples and generated counterfactual (CF) samples on SA, NLI, NER and RE tasks.}
\label{fig:llmcfgen_framework}
\end{figure*}

\paragraph{Spurious Correlation}
Let $y_+$ and $y_-$ denote the positive and negative sentiment polarity, respectively.
Assume in a collection of training samples, the context word $x_{t_1}$=\textit{``Nolan''} appears frequently in sentences with positive polarities and hardly ever in those with negative polarities.
Classical model training based on empirical risk minimization (ERM)~\cite{vapnik-1991-ERM}, indiscriminately learns from both spurious correlation $X_T \leftarrow C\rightarrow Y$ and causal correlation $X_C\rightarrow Y$.
Thus, we can obtain:
\begin{equation}
\small
    P(Y|X)=\frac{1}{||X_C \cup X_T||}\sum_{x_i \in (X_C \cup X_T) }P(Y|x_i),
\end{equation}
where $||X_C \cup X_T||$ denotes the number of words in the input $X$ and $Y=\{y_+, y_-\}$.
Due to the bias factor mentioned before, $P(y_+|x_{t_1})$ is much larger than $P(y_-|x_{t_1})$, and thus $P(Y|x_{t_1})$ tends to dominate the overall distribution.
That is, the model tends to learn $P(Y|X)$ from $x_{t_1}$ rather than $X_C$ during the training process.\looseness-1

\paragraph{Intervention and Counterfactual Generation}
To alleviate the issue above, one important way is to conduct causal intervention~\cite{pearl-1993-bayesian-intervention} via counterfactual generation.
Before performing causal intervention $do(X_T)$, one crucial step is to separate $X_T$ from $X_C$, i.e., causal words identification.
Next, we need to ensure that $X_T$ is unchanged, e.g., $do(X_T)=x_{t_1}$, and flip the sentiment polarity by changing $X_C$, e.g., \textit{``Nolan's films always \textbf{disappoint} me.''}.
After completing interventions for all samples, the counterfactual samples are augmented to the original samples so that both $P(y_+|do(X_T))$ and $P(y_-|do(X_T))$ are 1/2.
Hence $X_T$ has almost no contribution to $P(Y|X)$.
As shown in Fig.~\ref{fig:causal_formalism} (b), $P(Y|X)$ can be rewritten as:
\begin{equation}
\small
    P(Y|X)=\frac{1}{||X_C||}\sum_{x_i \in X_C}P(Y|x_i),
\end{equation}
which means that the model concentrates on learning $P(Y|X)$ from causal words $X_C$.

\subsection{LLMs for Counterfactual Generation}\label{sec:llm_cfgen}
As illustrated in Fig.~\ref{fig:llmcfgen_framework}, the proposed evaluation framework consists of three steps.

$S1$~(causal term identification): Separating causal words from context words. \looseness-1

$S2$~(label-controlled text infilling): Maintaining the context words unchanged, changing the label of the sample by altering the causal words. 

$S3$~(counterfactual data augmentation): Combining the original and counterfactual samples as training samples.

Since we aim to evaluate the counterfactual generation ability of LLMs, $S1$ and $S2$ are performed by prompting LLMs for text completion.
The combined samples after $S3$ are then used to train backbones for performing typical NLU tasks, e.g., SA.

\paragraph{Prompt Design}

For each labeled training sample, $x_{i}$, we construct a triplet prompt [$T^{p}$, $D^{p}$, $x_{i}^{p}$].
$T^{p}$ denotes task guidelines, including a precise definition of the task along with step-by-step descriptions on how to generate counterfactuals.
$D^{p}$ represents the demonstration part used to clarify the format of inputs and outputs.
$x_{i}^{p}$ denotes the standardized format of the original sample $x_{i}$, like that in the $D^{p}$. 
We provide such triplet prompts to LLMs, and expect LLMs to identify causal words, replace causal words and generate desired counterfactuals.



\subsection{Backbones for Data Augmentation}\label{sec:slm_backbones}
To measure the quality of the generated counterfactuals, we compare the performance of small language models (SLMs) trained with the original or counterfactually augmented data. 
We adopt SLMs like BERT~\cite{devlin2018bert} and BART~\cite{lewis2020bart} as backbones\footnote{We also experiment with LLMs as backbones for data augmentation. However, the time cost for testing is extremely expensive and the performance is not as good as that of SLMs. So we omit the evaluation on LLMs as backbones for data augmentation.}, which are typical for natural language understanding and generation tasks, respectively.
For BERT-based SLMs, the output embeddings of BERT are inputted to the MLP or CRF for further classification or tagging. The backbone is trained to minimize the cross-entropy loss.
For BART-based SLMs, the training goal of the model is to generate the target text following a pre-defined template and then we can de-linearize it into labels.

\section{Evalution of LLMs as the Counterfactual Generator}\label{sec:strengths_and_weaknesses}
In this section,  we choose GPT-3.5 as an example to evaluate and analyze the counterfactual generation ability of LLMs on four typical NLU tasks.

\begin{figure*}[htp]
\centering
\setlength{\abovecaptionskip}{0.1cm}
\includegraphics[width=1.0\textwidth]{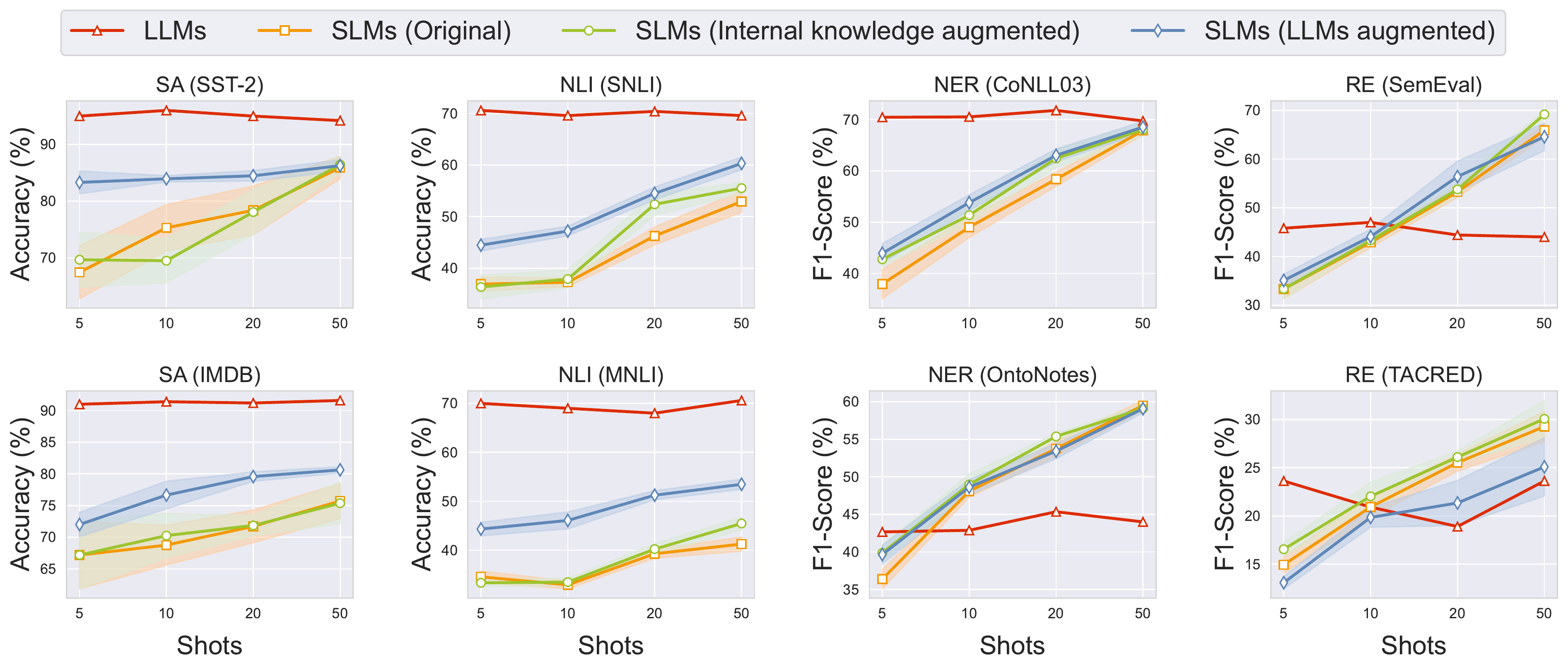}
\caption{Performance comparison under few-shot settings. The LLMs refer to GPT-3.5. The results of SLMs are obtained by averaging the performance of BERT-based and BART-based fine-tuned models.}
\label{fig:main_body_main_results_iid}
\end{figure*}

\subsection{Evaluation Protocal}\label{sec:evaluation_protocal}
\paragraph{Datasets and Evaluation Metrics}
We conduct experiments across various datasets.
Specifically, we adopt SST-2~\cite{socher-etal-2013-recursive} and IDMB~\cite{maas-EtAl:2011:ACL-HLT2011} for the SA task, SNLI~\cite{bowman-etal-2015-large} and MNLI~\cite{N18-1101} for the NLI task, CoNLL2003~\cite{tjong-kim-sang-de-meulder-2003-introduction} and OntoNotesV5~\cite{weischedel2013ontonotes} for the NER task, SemEval2010~\cite{hendrickx-etal-2010-semeval} and TACRED~\cite{zhang-etal-2017-position} for the RE task.
We use accuracy as the evaluation metric for SA and NLI and the micro-F1 for NER and RE.
We report mean accuracy or micro-F1 with standard deviation using 5 different seeds.

\paragraph{Few-shot Settings}
Spurious correlations are particularly prevalent in few-shot settings~\cite{nan-2021-uncovering-causalities}.
To evaluate the generated counterfactuals for mitigating such negative impact, we conduct experiments using randomly sampled \{5,10,20,50\}-shot training set on each dataset.
For the task where each sentence corresponds to a sample-label pair, i.e., SA, NLI, RE, we sample k samples for each class as the few-shot training set under the k-shot setting. 
For the task where each sentence corresponds to one or more sample-label pairs, i.e., the NER task, following~\cite{yang-katiyar-2020-structshot}, we adopt the greedy sampling algorithm\footnote{Note that it is unavoidable that the NER few-shot training set obtained using the greedy sampling algorithm may contain more than k samples corresponding to some classes.}.

\paragraph{Compared Methods for Typical NLU Tasks}
To investigate the efficacy of the generated counterfactuals by LLMs, we compare the performance of the following methods on typical NLU tasks.

{\textbf{LLMs}}: We test the performance of LLMs under few-shot settings as a comparative baseline.
Specifically, we adopt the widely used in-context learning approach, i.e., performing specific tasks by understanding instructions and in-context demonstrations in the prompt.
Generally, the prompt consists of three parts, i.e., [\{task definition\}, \{demonstrations\}, \{test sample\}].

{\textbf{SLMs (Original)}}: We test the original few-shot performance of SLMs via the BERT-based or BART-based fine-tuning methods.

{\textbf{SLMs (Internal knowledge augmented)}}: We augment the original SLMs with counterfactual data generated by internal knowledge tailored methods, including AutoCAD~\cite{wen-etal-2022-autocad} for SA and NLI tasks, CFGen~\cite{zeng2020cfgen} for NER, and CoCo~\cite{zhang2023coco} for RE.

{\textbf{SLMs (LLMs augmented)}}: We augment the original SLMs with counterfactual data generated by LLMs via the method introduced in Section~\ref{sec:llm_cfgen}. 
Notably, the purpose of including both SLMs (Internal knowledge augmented) and SLMs (LLMs augmented) here is to compare well-designed models with limited internal knowledge and general models with a large amount of external knowledge\footnote{Please refer to Appendix~\ref{sec:app:experimental_details} for more descriptions about experimental details.}.

\subsection{Discussion on Experimental Results}
Fig.~\ref{fig:main_body_main_results_iid} shows the experimental results\footnote{Please refer to Appendix~\ref{sec:app:detailed_main_results} for detailed results.} of various compared methods under few-shot settings, which we will discuss next.

\paragraph{Do SLMs Have Chances to Outperform LLMs?}
1) LLMs maintain clear advantages on relatively simple SA and NLI tasks, as well as on NER and RE tasks under extremely few-shot settings.
2) But for the NER and RE tasks, the advantage of LLMs seems to be not so obvious.
Normally, the increase in the number of labels requires an increase in task-specific knowledge.
However, the in-context learning approach may prevent LLMs from fully acquiring task-specific knowledge from the provided demonstrations. 
In other words, increasing the number of demonstrations may not notably improve the performance of LLMs~\cite{mayubo-2023-large-but}.
Therefore, for tasks with many labels, e.g., NER and RE\footnote{\small {There are 2, 3, \{4, 18\}, and \{10, 42\} labels in the SA, NLI, NER, and RE tasks, respectively.}}, the performance of fine-tuned SLMs consistently improves since SLMs can acquire more task-specific knowledge through fine-tuning while LLMs cannot.
Eventually, the performance of SLMs catches up or even surpasses that of LLMs.

\paragraph{Can LLMs-Generated Counterfactuals Enhance the Performance of SLMs?}
1) Counterfactual data generated by LLM significantly improve the performance of SLMs on SA and NLI tasks.
2) LLMs perform poorly on the more complex NER and RE tasks, where they only bring enhancements on some datasets (CoNLL2003 and SemEval), and even cause performance degradation on the TACRED (RE) dataset.
This is likely due to the failure of LLMs to consider entity constraints when generating counterfactuals, which will be analyzed later.

\paragraph{Can LLMs Always Achieve Better Augmentation Results than Internal Knowledge Tailored Methods?}
1) In most cases, LLMs demonstrate superior performance than internal knowledge tailored methods in generating counterfactuals, due to the vast inherent knowledge in them. 
2) Nevertheless, when engaged with the RE task, the internal knowledge tailored method CoCo is more effective. 
This is largely attributable to its meticulous design and the set constraints that guide the counterfactual generation process.

\subsection{Weaknesses Analysis of LLMs for Counterfactual Generation}\label{sec:llms_limitations}

\paragraph{The Quality of Generated Counterfactuals is Bounded by LLMs' Task-specific Performance}
We visualize the average performance of LLMs themselves and the average augmentation effects for SLMs on each dataset. 

\begin{figure}[h]
\vspace{-0mm}
\centering
\includegraphics[width=0.48\textwidth]{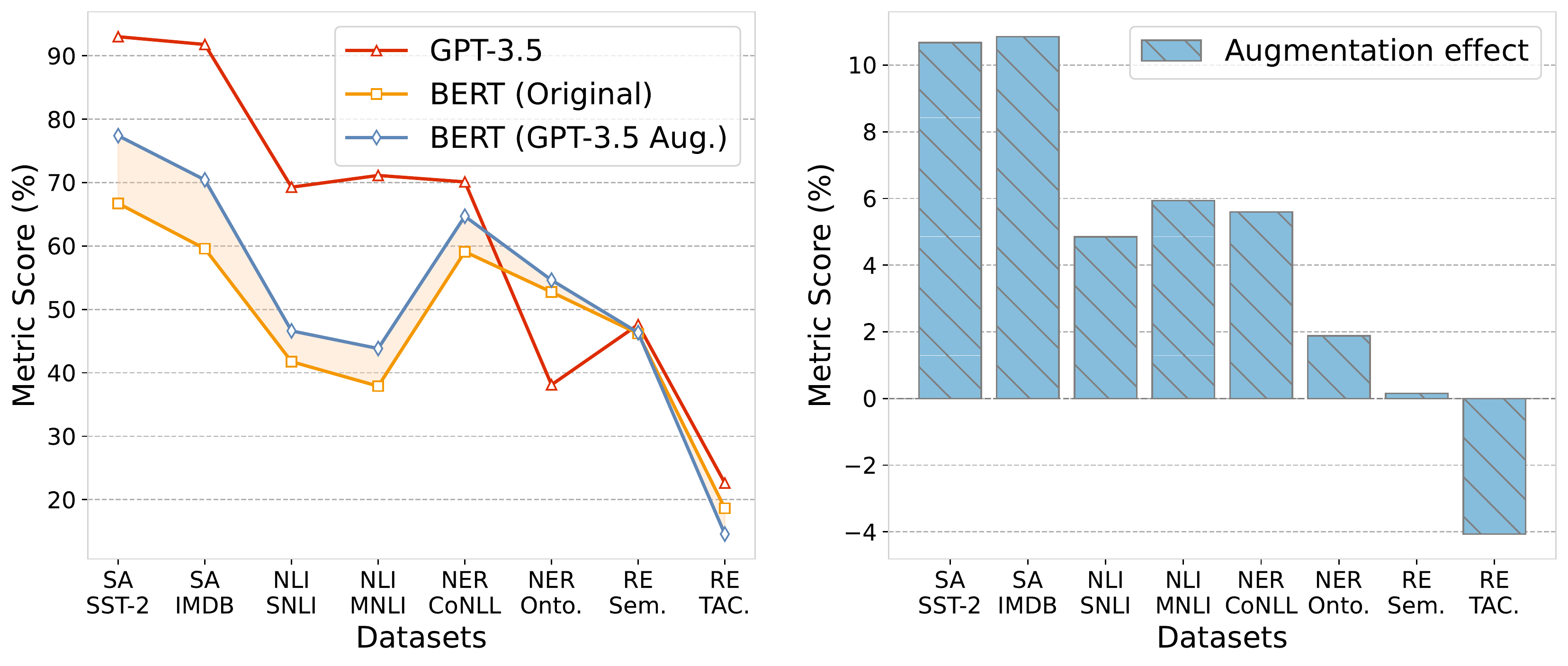}
\caption{Task-specific performance (left)  of LLMs and augmentation effects on SLMs (right).}
\label{fig:self_limitations}
\vspace{-0mm}
\end{figure}

As shown in the Fig.~\ref{fig:self_limitations}, we find a strong correlation between the counterfactual generation capability of LLMs and their task-specific performance.
Specifically, for the simpler SA task, LLMs can achieve up to about 93\% accuracy, and the generated counterfactual data have the most significant augmentation effect on SLMs, with about 11\% absolute increase.
On the TACRED dataset of the hard RE task, LLMs can only achieve a 22\% micro-F1 score. Correspondingly, the counterfactual data generated by LLMs have even a negative impact on the SLMs, i.e., a 4\% absolute decrease.
This finding indicates that the quality of generated counterfactuals is heavily bounded by the LLMs' task-specific performance, owing to the fact that we can only design prompts for counterfactual generation, which is far from expectations.

\begin{figure}[h]
\vspace{-0mm}
\centering
\includegraphics[width=0.48\textwidth]{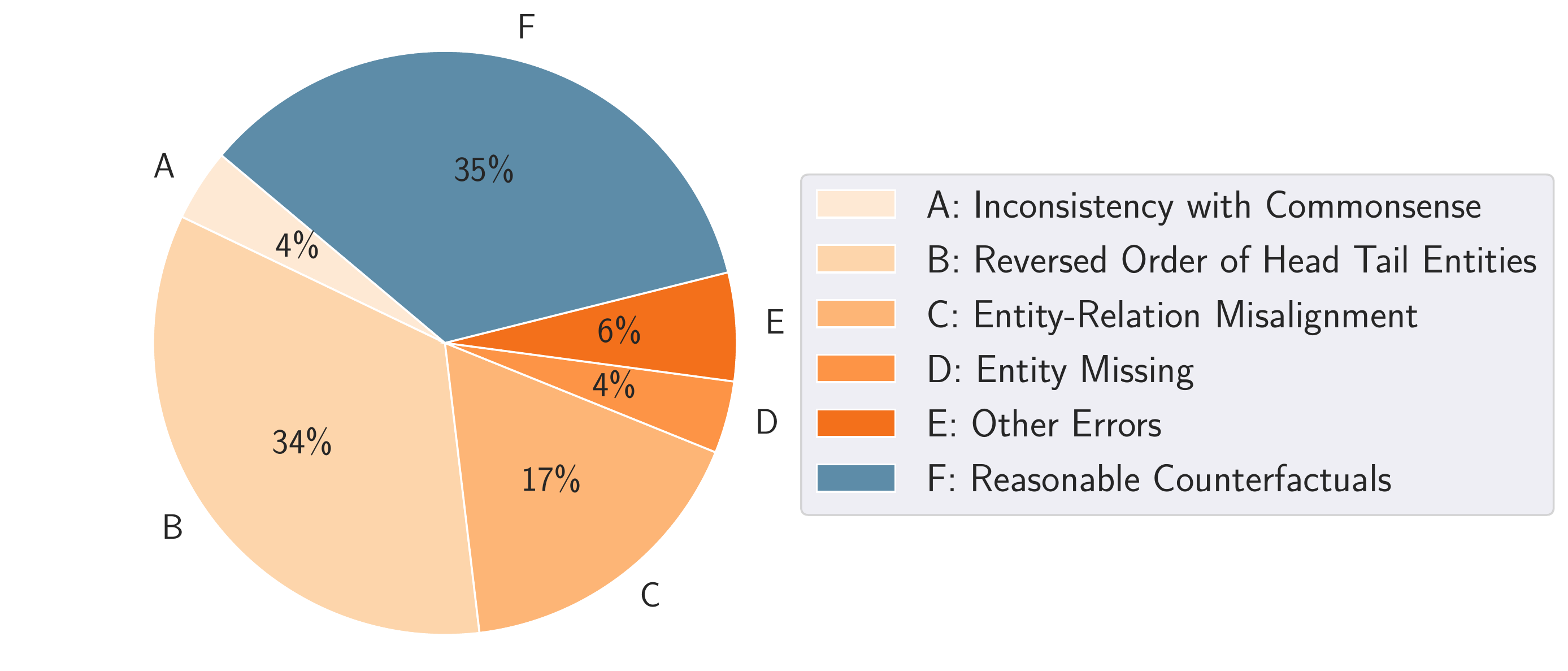}
\caption{Reasons that lead to unreasonable counterfactuals and corresponding proportions.}
\label{fig:human_evaluation_RE}
\vspace{-0mm}
\end{figure}

\paragraph{LLMs Fail to Fully Consider Entity Constraints when Generating Counterfactuals for RE}
In our previous experiments, we observe that for the RE task, the counterfactuals generated by GPT-3.5 might have a negative effect on SLMs.
To investigate this issue, we select 100 generated counterfactuals for human evaluation.
Specifically, we first determine whether the generated counterfactuals are reasonable, and then annotate the reasons for unreasonable ones.\looseness-1
The results are presented in Fig.~\ref{fig:human_evaluation_RE}.

\begin{table*}[t]
	\vspace{-0mm}
\centering
\setlength{\tabcolsep}{3mm}
\resizebox{\linewidth}{!}{
		\begin{tabular}{lll}
			\toprule
            \textbf{Case of Type B} & The \colorbox{blue1}{flight} departs from an \colorbox{blue1}{airport} on territory of a member state to which the Treaty applies. & Entity-Origin \\
            Counterfactual & The \colorbox{blue1}{flight} arrives at an \colorbox{blue1}{airport} on the territory of a member state to which the Treaty applies. & Destination-Entity \\
            
			\cmidrule(lr){1-1} \cmidrule(lr){2-2}\cmidrule(lr){3-3}
            \textbf{Case of Type C} & The woods that furnish the best \colorbox{blue1}{charcoal} for \colorbox{blue1}{painters} are the beech and vine. & Instrument-Agency\\
            Counterfactual & The beech and vine are the origin of the best \colorbox{blue1}{charcoal} for \colorbox{blue1}{painters}. & Entity-Origin\\
			\bottomrule
   
	   \end{tabular}
}
	\caption{Case study of noisy counterfactual samples generated by GPT-3.5 on the RE task. Cases are from the SemEval dataset. Entities are in \colorbox{blue1}{blue}.}
	\vspace{-0mm}
	\label{tab:case_study_noisy_samples}
\end{table*}

From Fig.~\ref{fig:human_evaluation_RE}, it can be seen that type B and type C, i.e., ``Reversed  Order of Head Tail Entities'' and ``Entity-Relation Misalignment'', are the two dominant causes of unreasonable counterfactuals.
We select two cases corresponding to these two types and present them in Table~\ref{tab:case_study_noisy_samples}.
In the case of type B, ``flight'' and ``airport'' should form the ``Entity-Destination'' relation, but not the reversed one, i.e., ``Destination-Entity''.
In the case of type C, the concerned entities are ``charcoal'' and ``painters''.
However, in the generated counterfactual sentence, the ``Origin'' part to form an ``Entity-Origin'' relation with the head entity ``charcoal'' is ``beech and vine'' rather than the tail entity ``painters''.
These findings indicate that LLMs still struggle with entity constraints such as the ``head-tail order'' and ``alignment with the counterfactual relation''.\looseness-1

\begin{figure}[h!]
\vspace{-0mm}
\centering
\includegraphics[width=0.48\textwidth]{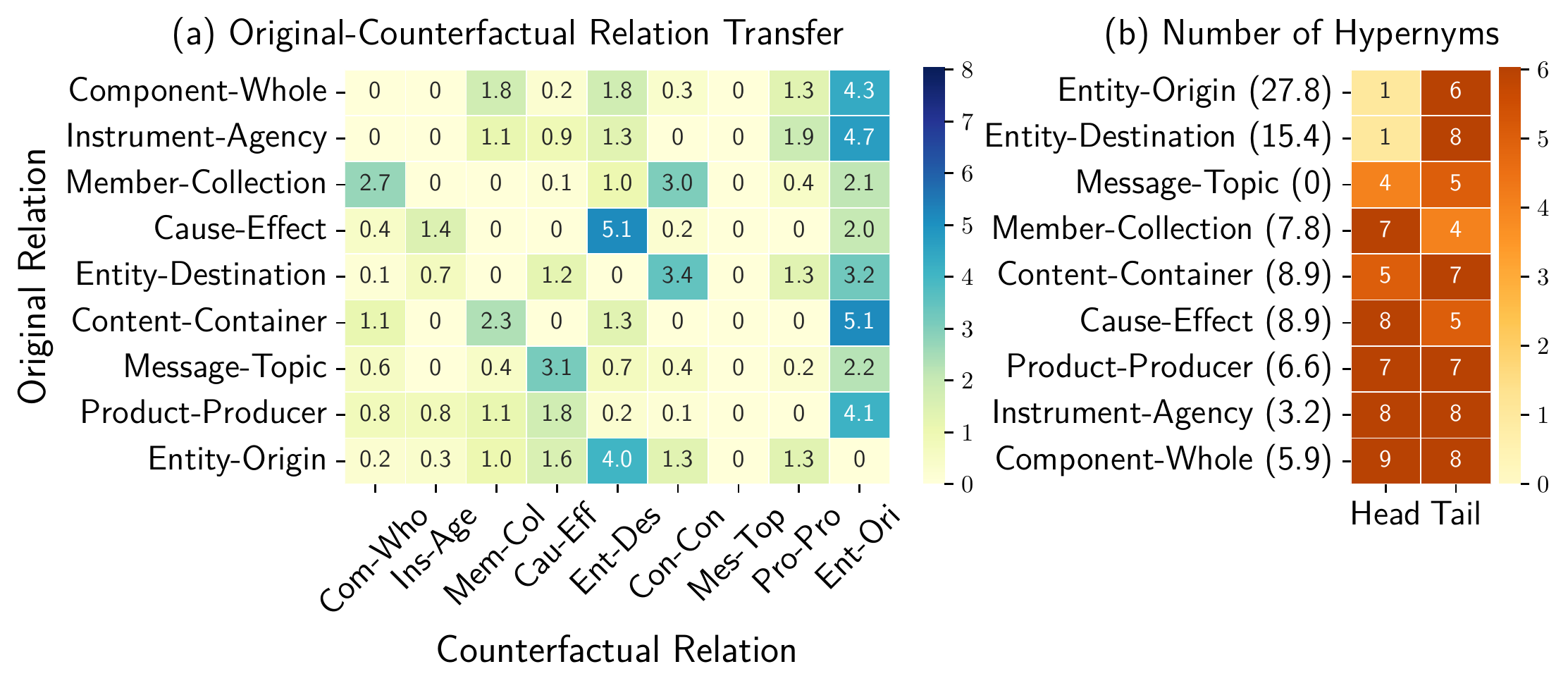}
\caption{(a) Visualization of original-counterfactual relation transfer frequency. The number represents the frequency of the corresponding transition every 100 samples. (b) Visualization of the number of hypernyms for each head and tail concept. The number in () represents the average frequency of being the target counterfactual relation for every 100 samples.}
\label{fig:relation_set_bias}
\vspace{-0mm}
\end{figure}

\paragraph{Selection Bias in LLMs Undermines Counterfactual Generation for the RE Task}
This section aims to investigate the potential selection bias in the choice of target counterfactual relations by LLMs.
Specifically, we select 100 samples, with 10 samples for each relation from the SemEval dataset, to observe the frequency of relation transfer. 
To exclude potential biases introduced by the demonstration in the prompt, we average the results using 9 different prompts with different target counterfactual relations in the demonstration. 
We then visualize the average frequency matrix of the ``original-counterfactual relation transfer'' for every 100 samples in Fig.~\ref{fig:relation_set_bias} (a).
We also adopt the hypernyms in WordNet~\cite{miller-1995-wordnet}, which can group similar words into a high-level concept\footnote{In WordNet, the fewer hypernyms a word has, the higher level of abstraction the word is.}, to observe such selection bias.
Fig.~\ref{fig:relation_set_bias} (b) shows the number of hypernyms of the head and tail concepts within each relation.



From Fig.~\ref{fig:relation_set_bias} (a), it is clear that LLMs tend to select certain relations as the target counterfactual ones.
Further, together with Fig.~\ref{fig:relation_set_bias} (b), we can see that, except for the ``Message-Topic'' relation, the frequency of a relation being chosen as a target counterfactual relation and the number of its hypernyms are negatively correlated.
In other words, LLMs prefer to choose more abstract relation types as the target counterfactual ones such as ``Entity-Origin''.
Such selection bias leads to a serious imbalance of labels in the generated counterfactual sample set, which may result in a performance decrease of the counterfactually augmented model.

\begin{table*}[htp]
\vspace{-0mm}
    \centering
    \setlength{\tabcolsep}{2.5mm}
    \resizebox{\linewidth}{!}{
        \begin{tabular}{lcccccccccc}
        \toprule
        \multicolumn{1}{c}{\multirow{3}{*}{\textbf{Aug. Method}}} & \multicolumn{4}{c}{\textbf{SA}}& \multicolumn{1}{c}{\multirow{3}{*}{\textbf{Avg.}}}                      & \multicolumn{4}{c}{\textbf{NLI}}       & \multicolumn{1}{c}{\multirow{3}{*}{\textbf{Avg.}}}             \\
        \cmidrule(lr){2-5}\cmidrule(lr){7-10}
        \multicolumn{1}{c}{}                                      & \multicolumn{2}{c}{SST-2} & \multicolumn{2}{c}{IMDB} && \multicolumn{2}{c}{SNLI} & \multicolumn{2}{c}{MNLI} \\
        \cmidrule(lr){2-3}\cmidrule(lr){4-5}\cmidrule(lr){7-8}\cmidrule(lr){9-10}
        \multicolumn{1}{c}{}                                      & 5-shot      & 10-shot     & 5-shot     & 10-shot&     & 5-shot     & 10-shot     & 5-shot     & 10-shot     \\
        \cmidrule(lr){1-1}\cmidrule(lr){2-5}\cmidrule(lr){6-6}\cmidrule(lr){7-10}\cmidrule(lr){11-11}
        None                                                    &59.67\small{\textpm5.39}&66.88\small{\textpm6.68}&57.05\small{\textpm1.40}&57.14\small{\textpm2.15}&  {60.19}  &37.67\small{\textpm1.29}&38.77\small{\textpm1.23}&35.03\small{\textpm1.72}&32.40\small{\textpm1.00}&{35.97}\\
        \cmidrule(lr){1-1}\cmidrule(lr){2-5}\cmidrule(lr){6-6}\cmidrule(lr){7-10}\cmidrule(lr){11-11}
        Llama2-7b                                               &{56.68\small{\textpm4.18}}&{65.45\small{\textpm6.73}}&\colorbox{orange1}{56.03\small{\textpm1.65}}&{54.77\small{\textpm1.76}}&  {58.23}  &\colorbox{orange1}{38.22\small{\textpm1.02}}&\colorbox{orange1}{43.39\small{\textpm1.73}}&{35.25\small{\textpm1.67}}&{32.45\small{\textpm1.02}}& {37.33}\\
        Llama2-7b-chat                                          &\colorbox{orange1}{81.88\small{\textpm1.70}}&\colorbox{orange1}{82.15\small{\textpm1.22}}&{52.88\small{\textpm0.82}}&\colorbox{orange1}{57.82\small{\textpm4.78}}&  \colorbox{orange1}{68.68}  &{37.88\small{\textpm4.27}}&{40.64\small{\textpm0.85}}&\colorbox{orange1}{41.24\small{\textpm1.81}}&\colorbox{orange1}{38.35\small{\textpm4.18}}&\colorbox{orange1}{39.53}\\
        \cmidrule(lr){1-1}\cmidrule(lr){2-5}\cmidrule(lr){6-6}\cmidrule(lr){7-10}\cmidrule(lr){11-11}
        Llama2-13b                                              &{56.57\small{\textpm5.31}}&{68.30\small{\textpm6.07}}&{52.55\small{\textpm1.38}}&{50.13\small{\textpm1.28}}& {56.89}   &\colorbox{orange1}{38.93\small{\textpm1.47}}&{39.67\small{\textpm1.45}}&{35.24\small{\textpm1.66}}&{34.23\small{\textpm1.47}}&{37.02}\\
        Llama2-13b-chat                                         &\colorbox{orange1}{81.86\small{\textpm1.97}}&\colorbox{orange1}{83.54\small{\textpm0.23}}&\colorbox{orange1}{54.41\small{\textpm3.05}}&\colorbox{orange1}{64.07\small{\textpm3.60}}&  \colorbox{orange1}{70.97}  &{34.03\small{\textpm1.25}}&\colorbox{orange1}{43.82\small{\textpm1.53}}&\colorbox{orange1}{36.23\small{\textpm1.90}}&\colorbox{orange1}{38.35\small{\textpm3.67}}&\colorbox{orange1}{38.11}\\
        \cmidrule(lr){1-1}\cmidrule(lr){2-5}\cmidrule(lr){6-6}\cmidrule(lr){7-10}\cmidrule(lr){11-11}
        Llama2-70b                                              &{65.14\small{\textpm6.10}}&{71.38\small{\textpm5.52}}&{53.93\small{\textpm1.57}}&{52.16\small{\textpm1.26}}&  {60.65}  &\colorbox{orange1}{41.00\small{\textpm1.83}}&\colorbox{orange1}{47.28\small{\textpm2.56}}&\colorbox{orange1}{37.87\small{\textpm1.35}}&{34.24\small{\textpm1.73}}&\colorbox{orange1}{40.10}\\
        Llama2-70b-chat                                         &\colorbox{orange1}{81.31\small{\textpm1.35}}&\colorbox{orange1}{84.04\small{\textpm0.44}}&\colorbox{orange1}{59.71\small{\textpm1.69}}&\colorbox{orange1}{62.96\small{\textpm3.41}}&  \colorbox{orange1}{72.00}  &{37.77\small{\textpm1.40}}&{44.50\small{\textpm0.88}}&{35.78\small{\textpm2.38}}&\colorbox{orange1}{38.60\small{\textpm3.11}}&{39.16}\\
        \bottomrule
        \end{tabular}
    }
\caption{Performance comparison of data augmentation experiments with counterfactuals generated by different LLMs. 
Here we use BERT-based SLMs. The better are in \colorbox{orange1}{orange}.}
\label{tab:compare_different_llms}
\vspace{-0mm}
\end{table*}

\paragraph{Promising Ways for Improving the Quality of Counterfactuals}
To address the two main causes of low-quality counterfactuals analyzed in previous paragraphs, i.e., inconsistency between entities and labels, and the selection bias of target counterfactual relations, there are three possible corresponding solutions.
1) {\textit{Consistency Filtering}}: We can employ SLMs trained on specific tasks to filter counterfactual samples for consistency. For example, only samples that SLMs correctly predict can be retained.
2) {\textit{Consistency Correction}}: We can utilize LLMs to conduct additional checks on the generated counterfactuals and correct them accordingly, such as whether the order of the head and tail entities is reversed, whether any entity is missing. 
3) {\textit{Correcting Selection Bias of LLMs}}: For example, we can restrict the choice of the target counterfactual relation based on commonsense knowledge related to the head and tail entities in advance. This prevents the LLMs from being influenced by their inherent bias when selecting the target counterfactual relation.

\section{Analysis on What Affects LLMs for Counterfactual Generation}
In this section, we will first analyze the impact of intrinsic properties of LLMs.
Then, we will analyze the impact of prompt designing.

\subsection{Intrinsic Properties of LLMs}\label{sec:llms_factors_pretrain}
To explore what intrinsic properties of LLMs affect the quality of counterfactual generation, we employ the Llama-2 series LLMs~\cite{touvron2023llama2} for counterfactual generation.
We mainly focus on two key factors, the parameter size of LLMs and whether using alignment techniques, e.g., reinforcement learning from human feedback (RLHF)~\cite{openai-2022-instructgpt}.
We choose the Llama-2 family of LLMs because there are six versions of Llama-2, i.e., \texttt{7b, 13b, 70b, 7b-chat, 13b-chat, 70b-chat}. 
The only difference between ``\texttt{\{7,13,70\}b}'' and ``\texttt{\{7,13,70\}b-chat}'' versions is that the latter adopt instruction-tuning and RLHF techniques for aligning with humans.
This provides good conditions for us to conduct controlled-variable experiments.
Please note that even using the powerful GPT-3.5, it is hard to generate high-quality counterfactuals for the NER and RE tasks, making them unsuitable for comparing the counterfactual generation capabilities of different LLMs.
Thus, we only select two relatively simple tasks, i.e., SA and NLI, for the experiments.

\begin{figure}[htp]
\vspace{-0mm}
\centering
\includegraphics[width=0.48\textwidth]{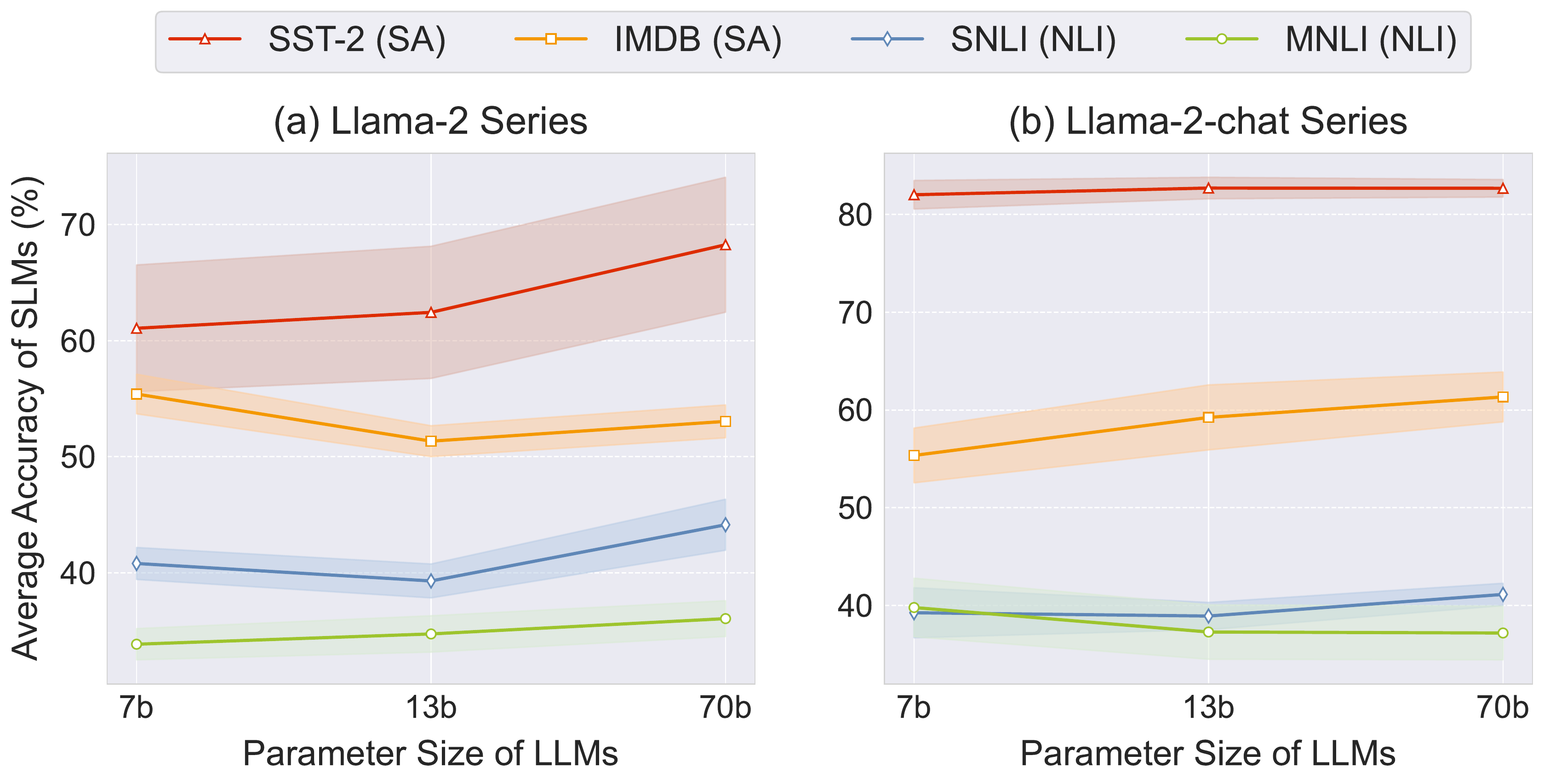}
\caption{Performance comparison of counterfactually augmented SLMs. The counterfactuals are generated by Llama-2 (left) and Llama-2-chat (right) series with different parameter sizes.
Results are obtained by averaging the performance of BERT-based SLMs under 5-shot and 10-shot settings.}
\label{fig:llama2_parameters_trend}
\end{figure}

\paragraph{Increasing Parameter Size cannot Improve Counterfactual Generation of LLMs}
To explore whether the parameter size of LLMs is critical to the counterfactual generation capability, 
we conduct counterfactual data augmentation experiments on SLMs, using counterfactuals generated by LLMs with different parameter sizes.
Fig.~\ref{fig:llama2_parameters_trend} illustrates the trend of counterfactually augmented SLMs performance with respect to the parameter size of LLMs.

It can be seen that despite the 10 times parameter size of LLMs (from 7b to 70b), the task performance of the counterfactually augmented SLMs is not significantly improved, i.e., the quality of generated counterfactuals is not notably improved.
This suggests that the counterfactual generation ability of LLMs does not improve as the number of model parameters rises, which is quite different from the widely held findings of previous studies for other tasks~\cite{wei-2022-LLMs-emergent}.

\begin{figure*}[h]
\vspace{-9mm}
\centering
\setlength{\abovecaptionskip}{0.1cm}
\includegraphics[width=1.0\textwidth]{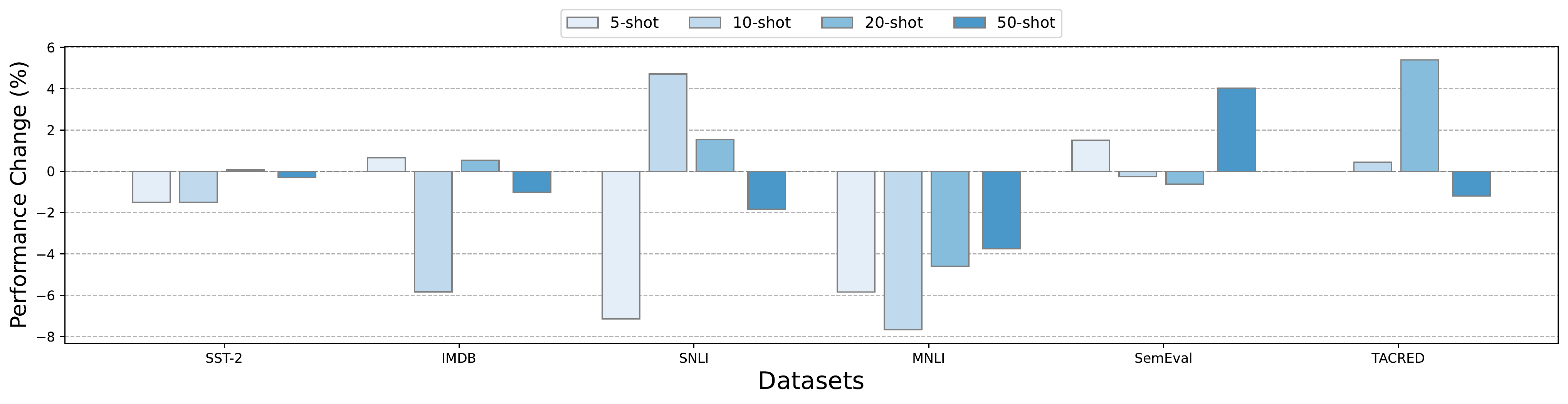}
\caption{The impact of chain-of-thought. The vertical axis represents the change in the performance after using chain-of-thought for counterfactual generation. Note that for the NER task, the causal words are the entities themselves. There is no need to find the causal words first, so it is not discussed here.}
\label{fig:iid_bert_CoT_bar_analysis}
\end{figure*}

\begin{table*}[htp!]
\centering
\setlength{\tabcolsep}{3mm}
\resizebox{\linewidth}{!}{
\begin{tabular}{lcccl}
\toprule
\textbf{Premise}                     & \multicolumn{4}{c}{The   echoes of colonialism are in the Railway Station, which is locally nicknamed   the Taj Mahal.}                                                                     \\
\midrule
\midrule

\multirow{6}{*}{\textbf{Hypothesis}} & \multicolumn{3}{c}{\textbf{Variant}}          & \textbf{Sentence}                                                                                                                                                      \\
\cmidrule(lr){2-4}\cmidrule(lr){5-5}
                                     & \multicolumn{3}{c}{Original}                                                                  & The Railway Station   is locally nicknamed the Taj hindra.                                                              \\
\cmidrule(lr){2-4}\cmidrule(lr){5-5}
                                     & \multicolumn{3}{c}{w/o   CoT}                                                                 & The Railway Station's   Moorish architecture is locally nicknamed the Taj Mahal.                                        \\
\cmidrule(lr){2-4}\cmidrule(lr){5-5}
                                     & \multirow{3}{*}{w/   CoT} & \multirow{2}{*}{Process} & \textit{Causal Identification} & \textit{The relationship "Contradiction" depends on "Taj   hindra" in the hypothesis sentence.}              \\
\cmidrule(lr){4-4}\cmidrule(lr){5-5}
                                     &                           &                          & \textit{Causal Replacement}      & \textit{To change the relation from "Contradiction" to "Entailment", "Taj hindra" is replaced by "Mahal".} \\
\cmidrule(lr){3-4}\cmidrule(lr){5-5}
                                     &                           & \multicolumn{2}{c}{Generated   Sentence}                          & The echoes of colonialism are in the Railway Station, which is locally nicknamed   the Mahal.                        \\
\bottomrule
\end{tabular}
}
\caption{An example of generated counterfactuals on the MNLI dataset without or with chain-of-thought (CoT). In this case, GPT-3.5  generates a counterfactual hypothesis sentence to change the relationship between the ``Premise'' and the ``Hypothesis'' sentence from ``Contradiction'' to ``Entailment''.}
	\vspace{-0mm}
	\label{tab:case_study_hallucination_samples}
\end{table*}

\paragraph{Alignment Techniques may Help Improving Counterfactual Generation of LLMs}
Table~\ref{tab:compare_different_llms} presents the performance comparison of SLMs on SA and NLI tasks, which are augmented using counterfactuals generated by Llama-2 and Llama-2-chat series.
For the SA task, the counterfactuals generated by Llama-2-chat series can bring a 10.4\%-14.1\% absolute accuracy increase averagely for SLMs than that of Llama-2 series.
Since the only difference between the Llama-2 and Llama-2-chat series is that the latter employs alignment techniques, i.e., instruction-tuning and RLHF, it is reasonable to conclude that the alignment techniques make important contributions to the improvement of the counterfactual generation capabilities.
But for the NLI task, the advantages of Llama-2-chat series over Llama-2 series are not significant.
Upon more detailed analysis, we observe that when generating ``Contradiction'' counterfactuals, Llama-2-chat series tend to favor generating ``Neutral'' samples by changing part of the semantics of the sentence, which introduces much noise.
This reflects the misalignment of label semantics understanding between LLMs and humans.
These findings inspire us to improve the counterfactual generation of LLMs by further exploring alignment techniques, e.g., 1) aligning LLMs' causal discovery process with that of humans to further enhance LLMs' causal discovery ability, 2) aligning LLMs' understanding of domain-specific label semantics with that of humans.

\subsection{Impact of Prompt Designing}\label{sec:llms_factors_prompt}
In this section, we adopt GPT-3.5 for exploring the impact of prompt designing\footnote{Please refer to Appendix~\ref{sec:app:detailed_main_results} for detailed results.}.
\paragraph{Task Guidelines are Critical for Counterfactual Generation}


To verify the necessity of providing detailed task guidelines in the prompt, we conduct comparison experiments of counterfactual generation with and without task guidelines.

\begin{table}[htp]
\vspace{-0mm}
    \centering
    \setlength{\tabcolsep}{3mm}
    \resizebox{\linewidth}{!}{
        \begin{tabular}{llccccc}
        \toprule
        \multirow{2}{*}{\textbf{Task}} & \multicolumn{1}{c}{\multirow{2}{*}{\textbf{Variant}}} & \multicolumn{4}{c}{\textbf{Shot}}                                  & \multicolumn{1}{c}{\multirow{2}{*}{\textbf{Avg.}}} \\
        \cmidrule(lr){3-6} 
                                       & \multicolumn{1}{c}{}                                  & 5-shot         & 10-shot        & 20-shot        & 50-shot        & \multicolumn{1}{c}{}                               \\
        \cmidrule(lr){1-2}\cmidrule(lr){3-6}\cmidrule(lr){7-7}
        \multirow{2}{*}{SA}            & w/ Task Guidelines                                          &\textbf{70.54}&\textbf{75.64}&\textbf{77.84}&\textbf{79.16}&\textbf{75.80}\\
                                       & w/o Task Guidelines                                         &69.67&75.37&76.23&78.71&75.00\\
        \cmidrule(lr){1-1}\cmidrule(lr){2-2}\cmidrule(lr){3-6} \cmidrule(lr){7-7}
        \multirow{2}{*}{NLI}           & w/ Task Guidelines                                          &\textbf{42.05}&44.91&\textbf{51.67}&\textbf{56.32}&\textbf{48.74}\\
                                       & w/o Task Guidelines                                         &39.05&\textbf{45.03}&51.52&54.05&47.41\\
        \cmidrule(lr){1-1}\cmidrule(lr){2-2}\cmidrule(lr){3-6} \cmidrule(lr){7-7}
        \multirow{2}{*}{NER}           & w/ Task Guidelines                                          &\textbf{50.78}&\textbf{60.66}&\textbf{68.09}&\textbf{72.55}&\textbf{63.02}\\
                                       & w/o Task Guidelines                                         &46.79&60.14&65.39&72.00&61.08\\
        \cmidrule(lr){1-1}\cmidrule(lr){2-2}\cmidrule(lr){3-6} \cmidrule(lr){7-7}
        \multirow{2}{*}{RE}            & w/ Task Guidelines                                          &18.84&28.38&35.74&41.44&31.10\\
                                       & w/o Task Guidelines                                         &\textbf{19.88}&\textbf{29.22}&\textbf{36.25}&\textbf{44.37}&\textbf{32.43}\\
        \bottomrule
        \end{tabular}

    }
\caption{The impact of task guidelines.}
\label{tab:bert_Instruct_analysis}
\end{table}

Table~\ref{tab:bert_Instruct_analysis} shows the results of the comparison experiments. 
It can be seen that the removal of the task guidelines leads to a performance drop on most datasets, e.g., a 2\% absolute decrease on the NER task.
The exception is the results of the RE task.
One possible reason for this particular case is that the counterfactuals generated by GPT-3.5 for the RE task have a negative impact on SLMs. 
When the task guidelines are removed, the number of generated counterfactuals decreases, thus reducing this negative effect.

\paragraph{Chain-of-thought does not Always Help}
Step-by-step task guidelines can help LLMs generate high-quality counterfactuals. 
Thus, intuitively, generating detailed explanations at each step may further improve the quality of generated counterfactuals.
To verify this assumption, we borrow the idea of chain-of-thought (CoT)~\cite{wei2022chain} and use GPT-3.5 to generate counterfactuals in a chain-like manner.
Specifically, we introduce two additional stages, ``Causal Words Identification'' and ``Causal Words Replacement'', and expect GPT-3.5 to generate explanations of these two steps before generating counterfactuals.

As shown in Fig.~\ref{fig:iid_bert_CoT_bar_analysis}, counter-intuitively, the counterfactuals generated by CoT do not lead to big improvements and even have significant decreases under some settings.
This might be due to the problem of inconsistency between the output and the process of counterfactual generation.
To have a close look, we select a representative case from the MNLI dataset and present it in Table~\ref{tab:case_study_hallucination_samples}.
It can be seen that, when using the prompt with CoT, the process of causal identification and replacement is correct.
However, the generated sentence is not the result of only replacing causal words in the original sentence, showing an inconsistent phenomenon.

\paragraph{Even unreasonable demonstration can yield reasonable counterfactuals}
Another question we are curious about is whether LLMs' ability to generate counterfactuals is acquired by learning the provided demonstration in the prompt, or from its large-scale pre-training process.
To answer this question, we replace the demonstration in the prompt with an unreasonable one for generating counterfactuals and show the results in Table~\ref{tab:bert_WrongDemo_bar_analysis}.\looseness=-1


\begin{table}[htp]
    \centering
    \setlength{\tabcolsep}{4mm}
    \resizebox{\linewidth}{!}{
        \begin{tabular}{cllllll}
        \toprule
        \multirow{2}{*}{\textbf{Task}} & \multicolumn{1}{c}{\multirow{2}{*}{\textbf{Variant}}} & \multicolumn{4}{c}{\textbf{Shot}}                                  & \multicolumn{1}{c}{\multirow{2}{*}{\textbf{Avg.}}} \\
        \cmidrule(lr){3-6} 
                                       & \multicolumn{1}{c}{}                                  & 5-shot         & 10-shot        & 20-shot        & 50-shot        & \multicolumn{1}{c}{}                               \\
        \cmidrule(lr){1-1}\cmidrule(lr){2-2}\cmidrule(lr){3-6} \cmidrule(lr){7-7}
        \multirow{2}{*}{SA}           & w/ Reasonable Demo                                            &\textbf{70.54}&\textbf{75.64}&\textbf{77.84}&\textbf{79.16}&\textbf{75.80}\\
                                       & w/ Unreasonable Demo                                         &69.50&73.47&76.40&78.66&74.51\\
        \cmidrule(lr){1-1}\cmidrule(lr){2-2}\cmidrule(lr){3-6} \cmidrule(lr){7-7}
        \multirow{2}{*}{NLI}           & w/ Reasonable Demo                                           &\textbf{42.05}&\textbf{44.91}&\textbf{51.67}&\textbf{56.32}&\textbf{48.74}\\
                                       & w/ Unreasonable Demo                                         &40.08&42.05&48.83&52.03&45.75\\

        \cmidrule(lr){1-1}\cmidrule(lr){2-2}\cmidrule(lr){3-6} \cmidrule(lr){7-7}
        \multirow{2}{*}{NER}            & w/ Reasonable Demo                                          &50.78&\textbf{60.66}&\textbf{68.09}&\textbf{72.55}&\textbf{63.02} \\
                                       & w/ Unreasonable Demo                                         &\textbf{52.59}&60.22&66.03&72.24&62.77\\
        \cmidrule(lr){1-2}\cmidrule(lr){3-6}\cmidrule(lr){7-7}
        \multirow{2}{*}{RE}            & w/ Reasonable Demo                                           &18.84&28.38&\textbf{35.74}&\textbf{41.44}&31.10\\
                                       & w/ Unreasonable Demo                                         &\textbf{19.71}&\textbf{28.86}&33.66&43.36&\textbf{31.40}\\
        \bottomrule
        \end{tabular}
    }
\caption{Results of comparison experiments with reasonable and unreasonable demonstrations.}
\label{tab:bert_WrongDemo_bar_analysis}
\vspace{-0mm}
\end{table}

From Table~\ref{tab:bert_WrongDemo_bar_analysis}, it is interesting that in most cases, the counterfactuals generated by GPT-3.5 using an unreasonable demonstration achieve comparable results to those by using a reasonable one.
This suggests that the demonstration in the prompt does not always teach LLMs the task goal of counterfactual generation, i.e., the counterfactual generation capability of LLMs is ``innate'' in them.

\section{Conclusion}
This paper presents the first evaluation framework and a systematical empirical study on the capability of LLMs in generating counterfactuals.
Experimental results on four typical NLU tasks including SA, NLI, NER, and RE demonstrate that LLMs can generate satisfactory counterfactuals in most cases. 
However, LLMs also have their weaknesses when dealing with complex tasks like RE due to the ignorance of entity constraints and inherent selection bias.
Notably, we also discover that alignment techniques are crucial for improving the counterfactual generation capabilities of LLMs.
This inspires us to explore alignment techniques for LLMs to generate high-quality counterfactuals in future work.

\section*{Acknowledgments}
This work was supported by the grant from the National Natural Science Foundation of China (NSFC) project (No. 62276193). It was also supported by the Joint Laboratory on Credit Science and Technology of CSCI-Wuhan University.

\section{Bibliographical References}\label{sec:reference}

\bibliographystyle{lrec-coling2024-natbib}
\bibliography{custom}

\clearpage

\begin{table*}[htp!]
    \centering
    \setlength{\tabcolsep}{1.8mm}
    \resizebox{\linewidth}{!}{
        \begin{tabular}{cccccccccc}
        \toprule
        \multirow{2}{*}{\textbf{Dataset}}      
        & \multirow{2}{*}{\textbf{Settings}}
        & \multirow{2}{*}{\textbf{\#Labels}} 
        & \multirow{2}{*}{\textbf{\#Train}} 
        & \multirow{1}{*}{\textbf{\#Internal}} 
        & \multirow{1}{*}{\textbf{\#GPT-3.5}} 
        & \multirow{1}{*}{\textbf{\#GPT-3.5}}  
        & \multirow{1}{*}{\textbf{\#GPT-3.5}}  
        & \multirow{1}{*}{\textbf{\#GPT-3.5}}
        & \multirow{2}{*}{\textbf{\#Test}}\\
        &&&& Aug. &  Aug. &(w/o Instruct.) Aug. &(w/ CoT) Aug. & (w/ Unreason. Demo) Aug.\\
        \cmidrule(lr){1-2}\cmidrule(lr){3-10}
        \multirow{4}{*}{SST-2}     & 5-shot   &\multirow{4}{*}{2}&   10            &3  &10  &10   &10                 &10     &\multirow{4}{*}{1,821} \\
                                   & 10-shot  &                  &   20            &4  &20  &20   &20                 &20     &             \\
                                   & 20-shot  &                  &   40            &14 &40  &40   &40                 &40     &              \\
                                   & 50-shot  &                  &   100           & 6 &100 &100  &100                 &100   &                \\
        \cmidrule(lr){1-1} \cmidrule(lr){2-2} \cmidrule(lr){3-10}
        \multirow{4}{*}{IMDB}      & 5-shot   &\multirow{4}{*}{2}&   10            & 0 &10 &10  &8 &9 &\multirow{4}{*}{25,000}  \\
                                   & 10-shot  &                  &   20            & 1 &20 &20  &19                  &19       &            \\
                                   & 20-shot  &                  &   40            & 1  &39 &40  &39                &39        &           \\
                                   & 50-shot  &                  &   100           & 7 &99 &99  &98                 &95        &           \\
        \midrule
        \midrule
        \multirow{4}{*}{SNLI}      & 5-shot   &\multirow{4}{*}{3}&   15            & 10 &26  &28   &30                &30       &\multirow{4}{*}{10,000} \\
                                   & 10-shot  &                  &   30            & 13 &58  &56   &60                &60       &                                  \\
                                   & 20-shot  &                  &   60            & 29 &120 &116  &120               &120      &                                     \\
                                   & 50-shot  &                  &   150           & 81 &300 &280  &300               &300      &                                      \\
        \cmidrule(lr){1-1} \cmidrule(lr){2-2} \cmidrule(lr){3-10}
        \multirow{4}{*}{MNLI}      & 5-shot   &\multirow{4}{*}{3}&   15            & 3  &30  &28   &30                 &30       &\multirow{4}{*}{10,000} \\
                                   & 10-shot  &                  &   30            & 9  &60  &60   &60                 &60       &                                  \\
                                   & 20-shot  &                  &   60            & 26 &120 &114  &120                &120      &                                       \\
                                   & 50-shot  &                  &   150           & 62 &300 &298  &300                &300      &                                      \\
        \midrule
        \midrule
        \multirow{4}{*}{CoNLL03}   & 5-shot   &\multirow{4}{*}{4}&   11            & 60  & 60  & 9   & \multirow{4}{*}{-}   & 42  &\multirow{4}{*}{3,453}                   \\
                                   & 10-shot  &                  &   22            & 174 & 166 & 28  &                      & 101 &                              \\
                                   & 20-shot  &                  &   39            & 291 & 285 & 70  &                      & 179 &                      \\
                                   & 50-shot  &                  &   96            & 707 & 673 & 189 &                      & 423 &                     \\
        \cmidrule(lr){1-1} \cmidrule(lr){2-2} \cmidrule(lr){3-10}
        \multirow{4}{*}{OntoNotes}   & 5-shot   &\multirow{4}{*}{18}&    54            & 524   & 309    & 132   & \multirow{4}{*}{-}   & 309     &\multirow{4}{*}{12,217}               \\
                                     & 10-shot  &                   &    123           & 1,088 & 1,155  & 316   &                      & 656     &                        \\
                                     & 20-shot  &                   &    209           & 1,997 & 2,073  & 481   &                      &  1,120  &                        \\
                                     & 50-shot  &                   &    603           & 5,191 & 5,463  & 1,265 &                      & 3,127   &                       \\
        \midrule
        \midrule
        \multirow{4}{*}{SemEval}   & 5-shot   &\multirow{4}{*}{10}&    50            & 0  &40  &29   &14                  &37  &\multirow{4}{*}{2,717}                  \\
                                   & 10-shot  &                   &    100           & 1  &71  &50   &25                  &70  &                                       \\
                                   & 20-shot  &                   &    200           & 5  &138 &93   &73                  &149 &                                   \\
                                   & 50-shot  &                   &    500           & 71 &370 &242  &147                 &372 &                                    \\
        \cmidrule(lr){1-1} \cmidrule(lr){2-2} \cmidrule(lr){3-10}
        \multirow{4}{*}{TACRED}    & 5-shot   &\multirow{4}{*}{41}&    210           & 1  &201   &175    &145                   &170     &\multirow{4}{*}{15,509}                \\
                                   & 10-shot  &                   &    416           & 10 &400   &349    &286                   &348     &                                    \\
                                   & 20-shot  &                   &    826           & 16 &775   &686    &572                   &671     &                                 \\
                                   & 50-shot  &                   &    1,994         & 90 &1,881 &1,652  &1,397                 &1,615   &                                     \\
        \bottomrule
        \end{tabular}}
    \caption{Statistics of datasets and the number of augmented samples for SA, NLI, NER and RE tasks by various counterfactual generation methods. 
    ``-'' denotes that there is no corresponding experiments. Because the causal words for the NER task are the entity words themselves. 
    There is no need to find the causal words first, so there is no need to discuss about the impact of CoT.
    }
    \label{tab:appendix_datasets}
\end{table*}


\appendix

\section{Experimental Details}\label{sec:app:experimental_details}

\subsection{Datasets}
Table~\ref{tab:appendix_datasets} shows statistics of various datasets used in our experiments and the number of augmented samples by various counterfactual generation methods.
Note that we select \texttt{mnli\_matched} as the test set for MNLI.

\subsection{Internal Augment Methods}
\textbf{AutoCAD}~\cite{wen-etal-2022-autocad} formulates the task of counterfactual generation as a label-controlled text infilling task. In other words, it aims to generate candidate words that can replace the trigger words in the original sentence, based on the provided target counterfactual label. Since there is no design involved to find the target counterfactual label in this method, it can only be used for counterfactual generation for SA and NLI where the target counterfactual labels are determined.\\
\textbf{CFGen}~\cite{zeng2020cfgen} employs substitution by using entity words from the same category present in the dataset to generate counterfactual samples.\\
\textbf{CoCo}~\cite{zhang2023coco} generates counterfactuals by flipping contextual words with the assistance of entity guidance. It leverages syntactic and semantic dependency graphs to identify linguistically substitutable spans from other contexts, while also flipping the labels associated with those spans.

\begin{table*}[]
    \centering
    \scriptsize  
    \begin{tabular}{cp{0.9\linewidth}}
        \toprule
       \textbf{Task}  &  \textbf{Templates} \\
       \midrule
       \multirow{2}{*}{SA}    &  \textbf{Input: } one long string of cliches   \\
       & \textbf{Output: } negative \\
       \midrule
       \multirow{2}{*}{NLI}   &  \textbf{Input: } premise: The new rights are nice enough hypothesis: Everyone really likes the newest benefits \\
       & \textbf{Output: } netural \\
       \midrule
       \multirow{3}{*}{RE}    &  \textbf{Input: } And then, when right-wing icon Barry Goldwater vacated his U.S. Senate seat in 1986, McCain vaulted into it. The relation between Barry Goldwater and U.S. Senate is \\
       & \textbf{Output: } employee of \\
       \midrule
       \multirow{2}{*}{NER}  & \textbf{Input: } For Mr. Sherwin, a conviction could carry penalties of five years in prison and a \$ 250,000 fine on each count \\
       & \textbf{Output: } person: Sherwin, date: five years, monetary: 250,000 \\
       \bottomrule
    \end{tabular}
    \caption{Templates of each task for BART-based SLMs.}
    \label{tab:templates}
\end{table*}

\subsection{Implement Details of LLMs}
We adopt the \texttt{gpt-3.5-turbo-0301} API from OpenAI for the experiments in Section~\ref{sec:strengths_and_weaknesses}.
For all experiments, the parameter temperature of LLMs is set to \texttt{0}.

\paragraph{LLMs for Specific Tasks}
In the prompt for specific tasks, i.e., SA, NLI, NER, and RE, we set the number of demonstrations as 2, 2, 6, 6, 4, 18, 10, and 16 for the SST-2, IMDB, SNLI, MNLI, CoNLL03, OntoNotes, SemEval, and TACRED datasets, respectively. The demonstrations are randomly selected from the labeled data.

\paragraph{LLMs for Counterfactual Generation}
In the prompt for generating counterfactuals, we provide a manually written demonstration.
Table \ref{prompt_sa_1}-\ref{prompt_re_3} shows the prompts used in the main comparison experiments and the analysis experiments for counterfactual generation of LLMs.

\subsection{Implement Details of SLMs}
\subsubsection{BERT-based}\label{appendix_bert_based_slm_implement_details}
BERT~\cite{devlin2018bert} equips multiple layers of Transformer encoder, which is trained by Masked Language Modeling and Next Sentence Prediction. BERT has achieved state-of-the-art results on a wide range of NLP tasks and has been widely adopted by researchers and industry practitioners. In this paper, we adopt the \emph{bert-base-uncased} as the SLM. Specifically, samples of different tasks are inputted to BERT as follow:
\begin{itemize}
    \item[$\bullet$]SA: [CLS] + \emph{sentence} + [SEP]
    \item[$\bullet$]NLI: [CLS] + \emph{premise sentence} + [SEP] + \emph{hypothesis sentence} + [SEP] 
    \item[$\bullet$]RE: [CLS] + \emph{entity$_1$} + [SEP] + \emph{entity$_2$} + [SEP] + \emph{sentence} + [SEP] 
    \item[$\bullet$]NER: [CLS] + \emph{sentence} + [SEP]
\end{itemize}
For SA, NLI, and RE tasks, the embeddings corresponding to [CLS] token are inputted to an MLP and Softmax layer to predict the final class of sentiment polarities of sentences, logical relationships of sentence-pairs, and entity-pairs relations, respectively. For the NER task, the embeddings corresponding to all tokens are inputted to a CRF layer to tag the position of entities. 

Furthermore, the learning rate, batch size, hidden size, and dropout rate of BERT-based backbones are set to 2 $\times$ $e^{-5}$, 64, 768, and 0.1, respectively. Notably, to ensure the reasonable of low-resource scenarios, we evaluate the backbones on test sets when their losses on train sets continuously increase over 3 epochs.

\subsubsection{BART-based}
With the continuous development of pre-trained generative models, many works try to convert various tasks into text generation tasks to utilize the label semantics better~\cite{zhang2021towards, lu2021text2event}. To verify the quality of counterfactual texts generated by LLMs more convincingly, we adopt the paradigm of text generation into the aforementioned tasks. We use BART-base~\cite{lewis2020bart} as the backbone, a sequence-to-sequence pre-trained model composed of a 6-layer encoder and a 6-layer decoder. Generally, the core of tackling tasks by text generation lies in designing specific templates. At the training stage, we need to convert the input and output of each sample into text-to-text form based on the pre-defined templates. At the inference stage, we need to de-linearize the label words from the target text generated by the model according to the pre-defined templates. The detailed templates are shown in Table~\ref{tab:templates}. Besides, we map the given label of RE and NER to the natural words to make the model better understand the label semantics as shown in Table~\ref{tab:label-word}.

Furthermore, batch size and learning rate are 64 and 5e-5, respectively. We evaluate the model on the training set per fixed steps \{50, 100, 200\}. When the performance on the training set reaches 1.0 or continuously decreases over three times, we will stop the model training and evaluate the model performance on the test set.

\section{Detailed Experimental Results}
\subsection{Results of Main Evaluation Experiments}\label{sec:app:detailed_main_results}
Tables \ref{tab:sa_main_results_bert}-\ref{tab:re_main_results_bert} show the detailed experimental results of the main comparison experiments.
We report the mean micro-F1 score with standard deviation using 5 different seeds.

\begin{table*}[t]
    \centering
    \setlength{\tabcolsep}{1.5mm}
    \resizebox{\linewidth}{!}{
        \begin{tabular}{llccccccccc}
        \toprule
        {\multirow{2}{*}{\textbf{Backbone}}} & {\multirow{2}{*}{\textbf{Aug. Method}}} & \multicolumn{4}{c}{\textbf{SST-2}} & \multicolumn{4}{c}{\textbf{IMDB}} \\
        \cmidrule(lr){3-6} \cmidrule(lr){7-10}
        \multicolumn{2}{c}{}                                 & 5-shot  & 10-shot  & 20-shot & 50-shot & 5-shot  & 10-shot & 20-shot & 50-shot \\
        \cmidrule(lr){1-2}\cmidrule(lr){3-6} \cmidrule(lr){7-10}
        \multirow{1}{*}{GPT-3.5}             & None     & 95.00 &  96.00  &  95.00  &  94.20  &  91.00  &  91.40  &  91.20  &  91.60  \\
        \cmidrule(lr){1-1}\cmidrule(lr){2-2}\cmidrule(lr){3-6} \cmidrule(lr){7-10}
        \multirow{3}{*}{BERT}      & None          &  59.67\small{\textpm5.39}  &  66.88\small{\textpm6.68}  &  73.25\small{\textpm6.33}  &  85.05\small{\textpm1.80}  &  57.05\small{\textpm1.40}  &  57.14\small{\textpm2.15}  &  57.14\small{\textpm2.48}  &  63.73\small{\textpm4.14}  \\
                                         & AutoCAD           &  62.26\small{\textpm6.04}  &  59.62\small{\textpm5.14}  &  72.33\small{\textpm5.53}  &  86.12\small{\textpm1.15}  &  57.05\small{\textpm1.40}  &  58.07\small{\textpm2.40}  &  57.02\small{\textpm1.95}  &  62.97\small{\textpm4.62}  \\
                                         & GPT-3.5         &  83.60\small{\textpm1.32}  &  84.29\small{\textpm0.25} &  85.58\small{\textpm0.49} &  85.98\small{\textpm0.60} &  57.48\small{\textpm2.40} &  66.98\small{\textpm2.00} &  70.09\small{\textpm0.80} &  72.33\small{\textpm0.25}  \\
        \cmidrule(lr){1-1}\cmidrule(lr){2-2}\cmidrule(lr){3-6} \cmidrule(lr){7-10}
        \multirow{3}{*}{BART}      
        & None &  75.21\small{\textpm3.94}  &  83.70\small{\textpm1.49}  &  83.50\small{\textpm2.25}  &  86.78\small{\textpm2.15}  &  77.32\small{\textpm9.26}  &  80.36\small{\textpm4.14}  &  86.29\small{\textpm2.67}  &  87.72\small{\textpm1.38}  \\
        & AutoCAD  &  77.06\small{\textpm3.63}  &  79.34\small{\textpm2.59}  &  83.81\small{\textpm2.36}  &  86.76\small{\textpm1.75}  &  77.32\small{\textpm9.26}  &  82.44\small{\textpm4.60}  &  86.70\small{\textpm1.26}  &  87.78\small{\textpm1.90}\\
        & GPT-3.5  &  82.99\small{\textpm2.67}  &  83.58\small{\textpm0.87} &  83.35\small{\textpm1.36} &  86.52\small{\textpm1.45} &  86.53\small{\textpm1.43} &  86.33\small{\textpm2.35} &  89.06\small{\textpm0.69} &  88.94\small{\textpm0.75}  \\
        \bottomrule
        \end{tabular}
    }
    \caption{Detailed results under few-shot settings of the SA task. Accuracy and standard deviations are reported.}
    \label{tab:sa_main_results_bert}
\end{table*}

\begin{table*}[t]
    \centering
    \setlength{\tabcolsep}{1.5mm}
    \resizebox{\linewidth}{!}{
        \begin{tabular}{llccccccccc}
        \toprule
        {\multirow{2}{*}{\textbf{Backbone}}} & {\multirow{2}{*}{\textbf{Aug. Method}}} & \multicolumn{4}{c}{\textbf{SNLI}} & \multicolumn{4}{c}{\textbf{MNLI}} \\
        \cmidrule(lr){3-6} \cmidrule(lr){7-10}
        \multicolumn{2}{c}{}                                 & 5-shot  & 10-shot  & 20-shot & 50-shot & 5-shot  & 10-shot & 20-shot & 50-shot \\
        \cmidrule(lr){1-2}\cmidrule(lr){3-6} \cmidrule(lr){7-10}
        \multirow{1}{*}{GPT-3.5}            & None     &  70.60  &  69.60  &  70.40  &  69.60  &  70.00  &  69.00  &  68.00  &  70.60  \\
        \cmidrule(lr){1-1}\cmidrule(lr){2-2}\cmidrule(lr){3-6} \cmidrule(lr){7-10}
        \multirow{3}{*}{BERT}      & None          &  37.67\small{\textpm1.29}  &  38.77\small{\textpm1.23}  & 51.09\small{\textpm1.75}  &  56.37\small{\textpm1.44}  &  35.03\small{\textpm1.72}  &  32.40\small{\textpm1.00}  & 41.67\small{\textpm1.31}  &  45.06\small{\textpm1.84}  \\
                                         & AutoCAD           &  35.61\small{\textpm2.00}  &  37.57\small{\textpm1.47}  &  53.57\small{\textpm1.27}  &  55.87\small{\textpm1.16}  &  33.53\small{\textpm1.16}  &  34.24\small{\textpm1.30}  &  41.40\small{\textpm1.25}  &  46.48\small{\textpm1.45}  \\
                                         & GPT-3.5         &  41.81\small{\textpm1.79}  &  45.11\small{\textpm0.69} &  52.52\small{\textpm1.45} &  59.91\small{\textpm0.96} &  42.28\small{\textpm1.58} &  44.72\small{\textpm1.94} &  50.83\small{\textpm0.94} &  52.73\small{\textpm1.23}  \\
        \cmidrule(lr){1-1}\cmidrule(lr){2-2}\cmidrule(lr){3-6} \cmidrule(lr){7-10}
        \multirow{3}{*}{BART}      
        & None      &  36.15\small{\textpm1.43}  &  35.74\small{\textpm0.48}  & 41.42\small{\textpm1.70}  &  49.47\small{\textpm2.76}  &  34.13\small{\textpm0.62}  &  33.47\small{\textpm0.60}  & 36.93\small{\textpm0.54}  &  37.47\small{\textpm1.04}  \\
        & AutoCAD  &  37.09\small{\textpm2.57}  &  38.19\small{\textpm1.96}  &  51.20\small{\textpm2.93}  &  55.17\small{\textpm0.93}  &  33.19\small{\textpm0.42}  &  32.76\small{\textpm0.23}  &  39.08\small{\textpm0.96}  &  44.48\small{\textpm1.18}  \\
        & GPT-3.5  &  47.10\small{\textpm0.48}  &  49.27\small{\textpm1.16} &  56.50\small{\textpm1.04} &  60.72\small{\textpm1.51} &  46.39\small{\textpm1.19} &  47.47\small{\textpm1.50} &  51.66\small{\textpm0.92} &  54.22\small{\textpm0.84}  \\
        \bottomrule
        \end{tabular}
    }
    \caption{Detailed results under few-shot settings of the NLI task. Accuracy and standard deviations are reported.}
    \label{tab:nli_main_results_bert}
\end{table*}

\begin{table*}[t]
    \centering
    \setlength{\tabcolsep}{1.5mm}
    \resizebox{\linewidth}{!}{
        \begin{tabular}{llccccccccc}
        \toprule
        {\multirow{2}{*}{\textbf{Backbone}}} & {\multirow{2}{*}{\textbf{Aug. Method}}} & \multicolumn{4}{c}{\textbf{CoNLL03}} & \multicolumn{4}{c}{\textbf{OntoNotes}} \\
        \cmidrule(lr){3-6} \cmidrule(lr){7-10}
        \multicolumn{2}{c}{}                                 & 5-shot  & 10-shot  & 20-shot & 50-shot & 5-shot  & 10-shot & 20-shot & 50-shot \\
        \cmidrule(lr){1-2}\cmidrule(lr){3-6} \cmidrule(lr){7-10}
        \multirow{1}{*}{GPT-3.5}             & None     &  70.46  &  70.54  &  71.80  &  69.75 &  42.67  &  42.88  &  45.35  &  44.00  \\
        \cmidrule(lr){1-1}\cmidrule(lr){2-2}\cmidrule(lr){3-6} \cmidrule(lr){7-10}
        \multirow{3}{*}{BERT}            & None              &  40.62\small{\textpm2.23}  &  56.53\small{\textpm1.81}  &  66.48\small{\textpm1.21}  &  75.91\small{\textpm0.83}  &  45.98\small{\textpm1.52}  &  58.11\small{\textpm0.83}  &  62.40\small{\textpm0.90}  &  68.03\small{\textpm1.03}  \\
                                         & CFGen             &  47.79\small{\textpm1.94}  &  58.50\small{\textpm0.92}  &  69.88\small{\textpm2.49}  &  76.18\small{\textpm1.55}  &  50.50\small{\textpm1.08}  &  58.49\small{\textpm1.22}  &  65.06\small{\textpm0.32}  &  67.37\small{\textpm0.82}  \\
                                         & GPT-3.5         &  49.13\small{\textpm2.17}  &  62.14\small{\textpm1.45}  &  73.00\small{\textpm0.66}  &  77.40\small{\textpm0.89}  &  52.42\small{\textpm1.59}  &  59.17\small{\textpm1.62}  &  63.19\small{\textpm0.98}  &  67.69\small{\textpm0.51}  \\
        \cmidrule(lr){1-1}\cmidrule(lr){2-2}\cmidrule(lr){3-6} \cmidrule(lr){7-10}
        \multirow{3}{*}{BART}      
        & None   &35.26\small{\textpm3.52}  &  41.50\small{\textpm2.20}  & 50.29\small{\textpm1.55}  &59.83\small{\textpm0.87}  &  26.79\small{\textpm1.06}  &  38.07\small{\textpm0.76}  &  45.10\small{\textpm1.09}  &  50.95\small{\textpm0.32}\\
        & CFGen &37.73\small{\textpm0.99}  &  44.23\small{\textpm2.00}  &  55.00\small{\textpm1.03}  &  59.93\small{\textpm1.01}  & 29.24\small{\textpm0.99}  &  39.47\small{\textpm1.54}  & 45.70\small{\textpm0.51}  &  50.80\small{\textpm0.25}\\ 
        & GPT-3.5 & 38.73\small{\textpm1.61}  &  45.47\small{\textpm1.63}  &  53.07\small{\textpm1.83}  & 59.70\small{\textpm1.35}  &  26.89\small{\textpm0.89}  &  38.07\small{\textpm0.42}  &  43.65\small{\textpm1.04}  &  50.44\small{\textpm0.58}\\ 
        \bottomrule
        \end{tabular}
    }
    \caption{Detailed results under few-shot settings of the NER task. micro-F1 scores and standard deviations are reported.}
    \label{tab:ner_main_results_bert}
\end{table*}

\begin{table*}[t]
    \centering
    \setlength{\tabcolsep}{1.5mm}
    \resizebox{\linewidth}{!}{
        \begin{tabular}{llccccccccc}
        \toprule
        {\multirow{2}{*}{\textbf{Backbone}}} & {\multirow{2}{*}{\textbf{Aug. Method}}} & \multicolumn{4}{c}{\textbf{SemEval}} & \multicolumn{4}{c}{\textbf{TACRED}} \\
        \cmidrule(lr){3-6} \cmidrule(lr){7-10}
        \multicolumn{2}{c}{}                                 & 5-shot  & 10-shot  & 20-shot & 50-shot & 5-shot  & 10-shot & 20-shot & 50-shot \\
        \cmidrule(lr){1-2}\cmidrule(lr){3-6} \cmidrule(lr){7-10}
        \multirow{1}{*}{GPT-3.5}             & None     &  45.80  &  47.00  &  44.40  &  44.00  &  23.62  &  20.91  &  18.90  &  23.62  \\
        \cmidrule(lr){1-1}\cmidrule(lr){2-2}\cmidrule(lr){3-6} \cmidrule(lr){7-10}
        \multirow{3}{*}{BERT}            & None               &  30.49\small{\textpm1.57}  &  {41.69\small{\textpm0.69}}  &  52.40\small{\textpm1.13}  &  64.06\small{\textpm1.18}  &  10.38\small{\textpm0.77}  &  17.17\small{\textpm1.12}  &  22.99\small{\textpm0.61}  &  27.69\small{\textpm1.99}  \\
                                         & CoCo               &  30.49\small{\textpm1.57}  &  {41.86\small{\textpm0.67}}  &  52.48\small{\textpm0.96}  &   68.52\small{\textpm0.86}  &   12.27\small{\textpm0.57}  &   18.60\small{\textpm2.03}  &   23.88\small{\textpm0.46}  &   29.82\small{\textpm2.75}  \\
                                         & GPT-3.5          &  28.89\small{\textpm1.22}  &  41.51\small{\textpm0.99} &  55.65\small{\textpm4.12} &  62.02\small{\textpm4.66} &  7.80\small{\textpm0.59} &  15.24\small{\textpm1.45} &  15.83\small{\textpm3.27} &  20.86\small{\textpm5.02}  \\
        \cmidrule(lr){1-1}\cmidrule(lr){2-2}\cmidrule(lr){3-6} \cmidrule(lr){7-10}
        \multirow{3}{*}{BART}      
        & None      &  36.14\small{\textpm2.45}  &  44.08\small{\textpm1.38} & 54.16\small{\textpm1.15}  &  67.80\small{\textpm1.57}  &  19.51\small{\textpm0.97}  &  24.69\small{\textpm1.11}  &  28.05\small{\textpm1.15}  & 30.82\small{\textpm1.13} \\
        & CoCo  &  36.14\small{\textpm2.45}  &  44.71\small{\textpm1.11} & 55.07\small{\textpm1.73}  &  69.88\small{\textpm0.75}  &  20.83\small{\textpm0.78}  &  25.46\small{\textpm0.89}  &  28.31\small{\textpm0.79}  &  30.29\small{\textpm1.06}\\
        & GPT-3.5  &  41.25\small{\textpm1.27}  &  46.62\small{\textpm2.09} & 57.16\small{\textpm2.23}  &  67.16\small{\textpm1.05}  &  18.36\small{\textpm0.67}  &  24.45\small{\textpm0.58}  &  26.85\small{\textpm1.41}  &  29.30\small{\textpm1.02}\\
        \bottomrule
        \end{tabular}
    }
    \caption{Detailed results under few-shot settings of the RE task. micro-F1 scores and standard deviations are reported.}
    \label{tab:re_main_results_bert}
\end{table*}

\subsection{Results of Analysis Experiments}\label{sec:app:detailed_analysis_results}
Tables \ref{tab:instruct_analysis_bert}-\ref{tab:wrong_demo_analysis_bert} show the detailed experimental results of the analysis experiments (Section~\ref{sec:llms_factors_prompt}).
We report the mean micro-F1 score with standard deviation using 5 different seeds.

\begin{table*}[htp!]
    \centering
    \setlength{\tabcolsep}{2mm}
    \resizebox{\linewidth}{!}{

        \begin{tabular}{llcccccccc}
        \toprule
        \multirow{2}{*}{\textbf{Shots}}   & \multirow{2}{*}{\textbf{Aug. Method}} & \multicolumn{2}{c}{\textbf{SA}}                     & \multicolumn{2}{c}{\textbf{NLI}}                    & \multicolumn{2}{c}{\textbf{NER}}           & \multicolumn{2}{c}{\textbf{RE}}                    \\
        \cmidrule(lr){3-4} \cmidrule(lr){5-6} \cmidrule(lr){7-8}\cmidrule(lr){9-10}
                             &                                  & SST-2                    & IMDB                     & SNLI                     & MNLI                     & CoNLL03                  & OnotoNotes           & SemEval                  & TACRED              \\
        \cmidrule(lr){1-2} \cmidrule(lr){3-10}                                  
        \multirow{2}{*}{\textbf{5-shot}}  & \textbf{GPT-3.5 w/ Instructions}                            & 83.60\small{\textpm1.32} & 57.48\small{\textpm2.40} & 41.81\small{\textpm1.79} & 42.28\small{\textpm1.58}  & 49.13\small{\textpm2.17} & 52.42\small{\textpm1.59}  & 29.89\small{\textpm1.22} & 7.80\small{\textpm0.59} \\
                                                                        & \textbf{GPT-3.5 w/o Instructions}         & 82.86\small{\textpm1.91} & 56.49\small{\textpm3.14} & 40.85\small{\textpm1.40} & 37.25\small{\textpm1.58}  & 45.39\small{\textpm2.05} & 48.19\small{\textpm0.99}  & 30.70\small{\textpm1.57} & 9.05\small{\textpm0.70} \\
                                                                        
        \cmidrule(lr){1-1} \cmidrule(lr){2-2} \cmidrule(lr){3-10} 
                                      \multirow{2}{*}{\textbf{10-shot}} & \textbf{GPT-3.5 w/ Instructions}    & 84.29\small{\textpm0.25} & 66.98\small{\textpm2.00} & 45.11\small{\textpm0.69} & 44.72\small{\textpm1.94}  & 62.14\small{\textpm1.45} & 59.17\small{\textpm1.62}& 41.51\small{\textpm0.99} & 15.24\small{\textpm1.45}  \\
                                                                        & \textbf{GPT-3.5 w/o Instructions}        &84.63\small{\textpm0.24} & 66.11\small{\textpm2.08} & 47.17\small{\textpm0.73} & 42.89\small{\textpm2.75}   & 61.32\small{\textpm1.05} & 58.96\small{\textpm1.25}  & 41.77\small{\textpm0.66} &16.66\small{\textpm1.41} \\
        \cmidrule(lr){1-1} \cmidrule(lr){2-2} \cmidrule(lr){3-10} 
                                      \multirow{2}{*}{\textbf{20-shot}} & \textbf{GPT-3.5 w/ Instructions}     & 85.58\small{\textpm0.49} & 70.09\small{\textpm0.80} & 52.52\small{\textpm1.45} & 50.83\small{\textpm0.94}  & 73.00\small{\textpm0.66} & 63.19\small{\textpm0.98} & 55.65\small{\textpm4.12} & 15.83\small{\textpm3.27}        \\
                                                                        & \textbf{GPT-3.5 w/o Instructions}         & 85.21\small{\textpm0.48} & 67.25\small{\textpm1.97} & 50.99\small{\textpm0.72} & 52.04\small{\textpm0.75}  & 68.73\small{\textpm1.54} & 62.05\small{\textpm1.06}  & 49.72\small{\textpm4.56} & 22.79\small{\textpm1.49} \\
        \cmidrule(lr){1-1} \cmidrule(lr){2-2} \cmidrule(lr){3-10} 
                                      \multirow{2}{*}{\textbf{50-shot}} & \textbf{GPT-3.5 w/ Instructions}    & 85.98\small{\textpm0.60} & 72.33\small{\textpm0.25} & 59.91\small{\textpm0.96} & 52.73\small{\textpm1.23}  & 77.40\small{\textpm0.89} & 67.69\small{\textpm0.51} & 62.02\small{\textpm4.66} & 20.86\small{\textpm5.02}   \\
                                                                        & \textbf{GPT-3.5 w/o Instructions}         & 85.68\small{\textpm1.46} & 71.73\small{\textpm0.57} & 56.62\small{\textpm1.53} & 51.47\small{\textpm0.98}  & 76.17\small{\textpm0.49} & 67.82\small{\textpm0.72}  & 64.91\small{\textpm0.40} & 23.83\small{\textpm4.10} \\
        \bottomrule
        \end{tabular}
    }
    \caption{Detailed results under few-shot settings for the analysis on the impact of removing instructions. Accuracy or micro-F1 scores and standard deviations are reported. Here BERT is used as the SLMs backbone.}
    \label{tab:instruct_analysis_bert}
\end{table*}

\begin{table*}[htp!]
    \centering
    \setlength{\tabcolsep}{5mm}
    \resizebox{\linewidth}{!}{

        \begin{tabular}{llcccccc}
        \toprule
        \multirow{2}{*}{\textbf{Shots}}   & \multirow{2}{*}{\textbf{Aug. Method}} & \multicolumn{2}{c}{\textbf{SA}}                     & \multicolumn{2}{c}{\textbf{NLI}}                    & \multicolumn{2}{c}{\textbf{RE}}                             \\
        \cmidrule(lr){3-4} \cmidrule(lr){5-6} \cmidrule(lr){7-8}
                                       &                                  & SST-2                    & IMDB                     & SNLI                     & MNLI                     & SemEval                  & TACRED                      \\
        \cmidrule(lr){1-2} \cmidrule(lr){3-8}                                  
        \multirow{3}{*}{\textbf{5-shot}}  
                                                                         & \textbf{GPT-3.5 w/o CoT}        & 83.60\small{\textpm1.32} & 57.48\small{\textpm2.40} & 41.81\small{\textpm1.79} & 42.28\small{\textpm1.58} & 29.89\small{\textpm1.22} & 7.80\small{\textpm0.59} \\
                                                                        & \textbf{GPT-3.5 w/ CoT}       & 82.10\small{\textpm2.44} & 58.14\small{\textpm3.31} & 34.67\small{\textpm1.91} & 36.44\small{\textpm1.36} & 31.40\small{\textpm1.81} & 7.79\small{\textpm0.44} \\
                                                                        
        \cmidrule(lr){1-1} \cmidrule(lr){2-2} \cmidrule(lr){3-8} 
                                      \multirow{3}{*}{\textbf{10-shot}} & \textbf{GPT-3.5 w/o CoT}       & 84.29\small{\textpm0.25} & 66.98\small{\textpm2.00} & 45.11\small{\textpm0.69} & 44.72\small{\textpm1.94} & 41.51\small{\textpm0.99} & 15.24\small{\textpm1.45}  \\
                                                                        & \textbf{GPT-3.5 w/ CoT}       & 82.80\small{\textpm0.95} & 61.15\small{\textpm5.05} & 49.82\small{\textpm1.66} & 37.05\small{\textpm1.59} & 41.26\small{\textpm0.91} & 15.68\small{\textpm1.69} \\
        \cmidrule(lr){1-1} \cmidrule(lr){2-2} \cmidrule(lr){3-8} 
                                      \multirow{3}{*}{\textbf{20-shot}} & \textbf{GPT-3.5 w/o CoT} & 85.58\small{\textpm0.49} & 70.09\small{\textpm0.80} & 52.52\small{\textpm1.45} & 50.83\small{\textpm0.94} & 55.65\small{\textpm4.12} & 15.83\small{\textpm3.27}        \\
                                                                        & \textbf{GPT-3.5 w/ CoT}      & 85.64\small{\textpm0.59} & 70.63\small{\textpm0.89} & 54.05\small{\textpm2.07} & 46.23\small{\textpm1.70} & 55.02\small{\textpm1.00} & 21.22\small{\textpm0.79} \\
        \cmidrule(lr){1-1} \cmidrule(lr){2-2} \cmidrule(lr){3-8} 
                                      \multirow{3}{*}{\textbf{50-shot}} & \textbf{GPT-3.5 w/o CoT}    & 85.98\small{\textpm0.60} & 72.33\small{\textpm0.25} & 59.91\small{\textpm0.96} & 52.73\small{\textpm1.23} & 62.02\small{\textpm4.66} & 20.86\small{\textpm5.02}   \\
                                                                        & \textbf{GPT-3.5 w/ CoT}         & 85.68\small{\textpm0.62} & 71.32\small{\textpm0.93} & 58.08\small{\textpm1.96} & 48.99\small{\textpm0.63} & 66.05\small{\textpm0.47} & 19.66\small{\textpm0.90} \\
        \bottomrule
        \end{tabular}
    }
    \caption{Detailed results under few-shot settings for the analysis on the impact of using chain-of-thought. Accuracy or micro-F1 scores and standard deviations are reported. Here BERT is used as the SLMs backbone.}
    \label{tab:cot_analysis_bert}
\end{table*}

\begin{table*}[htp!]
    \centering
    \setlength{\tabcolsep}{2mm}
    \resizebox{\linewidth}{!}{

        \begin{tabular}{llcccccccc}
        \toprule
        \multirow{2}{*}{\textbf{Shots}}   & \multirow{2}{*}{\textbf{Aug. Method}} & \multicolumn{2}{c}{\textbf{SA}}                     & \multicolumn{2}{c}{\textbf{NLI}}                    & \multicolumn{2}{c}{\textbf{NER}}           & \multicolumn{2}{c}{\textbf{RE}}                    \\
        \cmidrule(lr){3-4} \cmidrule(lr){5-6} \cmidrule(lr){7-8}\cmidrule(lr){9-10}
                                    &                                  & SST-2                    & IMDB                     & SNLI                     & MNLI                     & CoNLL03                  & OnotoNotes           & SemEval                  & TACRED              \\
        \cmidrule(lr){1-2} \cmidrule(lr){3-10}                                  
        \multirow{2}{*}{\textbf{5-shot}}  & \textbf{GPT-3.5 w/ Reason. Demo}         & 83.60\small{\textpm1.32} & 57.48\small{\textpm2.40} & 41.81\small{\textpm1.79} & 42.28\small{\textpm1.58}  & 49.13\small{\textpm2.17} & 52.42\small{\textpm1.59}  & 29.89\small{\textpm1.22} & 7.80\small{\textpm0.59} \\
                                                                        & \textbf{GPT-3.5 w/ Unreason. Demo}          & 82.15\small{\textpm1.57} & 56.85\small{\textpm3.90} & 39.61\small{\textpm1.54} & 40.54\small{\textpm2.08}  & 52.80\small{\textpm1.00} & 52.38\small{\textpm0.83} & 31.27\small{\textpm1.80} & 8.16\small{\textpm0.85} \\
                                                                        
        \cmidrule(lr){1-1} \cmidrule(lr){2-2} \cmidrule(lr){3-10} 
                                      \multirow{2}{*}{\textbf{10-shot}} & \textbf{GPT-3.5 w/ Reason. Demo}          & 84.29\small{\textpm0.25} & 66.98\small{\textpm2.00} & 45.11\small{\textpm0.69} & 44.72\small{\textpm1.94}  & 62.14\small{\textpm1.45} & 59.17\small{\textpm1.62}& 41.51\small{\textpm0.99} & 15.24\small{\textpm1.45}  \\
                                                                        & \textbf{GPT-3.5 w/ Unreason. Demo}           & 83.66\small{\textpm1.05} & 63.28\small{\textpm2.36} & 38.40\small{\textpm1.98} & 45.69\small{\textpm1.83}  & 60.95\small{\textpm2.83 } & 59.48\small{\textpm1.63 } & 40.18\small{\textpm0.57} & 17.54\small{\textpm2.37} \\
        \cmidrule(lr){1-1} \cmidrule(lr){2-2} \cmidrule(lr){3-10} 
                                      \multirow{2}{*}{\textbf{20-shot}} & \textbf{GPT-3.5 w/ Reason. Demo}          & 85.58\small{\textpm0.49} & 70.09\small{\textpm0.80} & 52.52\small{\textpm1.45} & 50.83\small{\textpm0.94}  & 73.00\small{\textpm0.66} & 63.19\small{\textpm0.98} & 55.65\small{\textpm4.12} & 15.83\small{\textpm3.27}        \\
                                                                        & \textbf{GPT-3.5 w/ Unreason. Demo}           & 85.34\small{\textpm0.62} & 67.45\small{\textpm0.73} & 46.91\small{\textpm1.86} & 50.75\small{\textpm1.52 } & 69.11\small{\textpm4.06 } & 62.96\small{\textpm1.1 } & 49.94\small{\textpm1.02} & 17.39\small{\textpm1.52} \\
        \cmidrule(lr){1-1} \cmidrule(lr){2-2} \cmidrule(lr){3-10} 
                                      \multirow{2}{*}{\textbf{50-shot}} & \textbf{GPT-3.5 w/ Reason. Demo}          & 85.98\small{\textpm0.60} & 72.33\small{\textpm0.25} & 59.91\small{\textpm0.96} & 52.73\small{\textpm1.23}  & 77.40\small{\textpm0.89} & 67.69\small{\textpm0.51} & 62.02\small{\textpm4.66} & 20.86\small{\textpm5.02}   \\
                                                                        & \textbf{GPT-3.5 w/ Unreason. Demo}           & 84.81\small{\textpm0.38} & 72.51\small{\textpm0.44} & 51.86\small{\textpm1.59} & 52.20\small{\textpm0.95 } & 77.32\small{\textpm0.66 } & 67.16\small{\textpm0.42 } & 62.38\small{\textpm0.45} & 24.33\small{\textpm3.64} \\
        \bottomrule
        \end{tabular}
    }
    \caption{Detailed results under few-shot settings for the analysis on the impact of providing unreasonable demonstration. Accuracy or micro-F1 scores and standard deviations are reported. Here BERT is used as the SLMs backbone.}
    \label{tab:wrong_demo_analysis_bert}
\end{table*}

\begin{table*}[]
\resizebox{\linewidth}{!}{
    \begin{tabular}{l}
        \toprule
        \begin{tabular}[c]{@{}l@{}}
        Task Definition: Revise a given sentence with minimal changes to alter its sentiment polarity.\\ 
        Instruction: This process consists of two steps. The first step is to identify the words in the given sentence that have the highest \\ 
        potential to change the sentiment polarity after substitution, known as the causal words.  The second step is to select appropriate \\ 
        replacement words for the causal words that will change the sentiment polarity of the sentence to the desired polarity.\\ 
        Demonstration:\\ 
        Given Sentence: "The movie is the best that I have ever seen."\\ 
        Current Sentiment Polarity: "positive"\\ 
        Target Sentiment Polarity: "negative"\\ 
        Revised Sentence: "The movie is the baddest that I have ever seen."\\ 
        Based on the given task definition and instruction, complete the following text by imitating the given demonstration.\\ 
        Given Sentence: "but it also has many of the things that made the first one charming ."\\ 
        Current Sentiment Polarity: "positive"\\ 
        Target Sentiment Polarity: "negative"\end{tabular} \\ 
        \bottomrule
    \end{tabular}
}
\caption{Prompts for counterfactual generation for the SA task.}
\label{prompt_sa_1}
\end{table*}

\begin{table*}[]
\resizebox{\linewidth}{!}{
    \begin{tabular}{l}
        \toprule
        \begin{tabular}[c]{@{}l@{}}
        Task Definition: Revise a given sentence with minimal changes to alter its sentiment polarity.\\ 
        Instruction: This process consists of two steps. The first step is to identify the words in the given sentence that have the highest \\
        potential to change the sentiment polarity after substitution, known as the causal words. The second step is to select appropriate \\
        replacement words for the causal words that will change the sentiment polarity of the sentence to the desired polarity.\\ Demonstration:\\ Given Sentence: "The movie is the best that I have ever seen."\\ Current Sentiment Polarity: "positive"\\ Causal Words Identification: The sentiment polarity "positive" depends on words "best".\\ Target Sentiment Polarity: "negative"\\ 
        Causal Words Replacement: To change the sentiment polarity of the given sentence from "positive" \\
        to "negative", causal words "best" are replaced by "baddest".\\ 
        Revised Sentence: "The movie is the baddest that I have ever seen."\\ Based on the given task definition and instruction, complete the following text by imitating the given demonstration.\\ Given Sentence: "This movie could not satisfy you."\\ Current Sentiment Polarity: "negative"\\ Causal Words Identification: \\ Target Sentiment Polarity: "positive"\\ Causal Words Replacement: \\ Revised Sentence:\end{tabular} \\ 
        \bottomrule
    \end{tabular}
}
\caption{Prompts for counterfactual generation for the SA task (with chain-of-thought).}
\end{table*}

\begin{table*}[]
\resizebox{\linewidth}{!}{
    \begin{tabular}{l}
        \toprule
        \begin{tabular}[c]{@{}l@{}}
        Task Definition: Revise a given sentence with minimal changes to alter its sentiment polarity.\\ 
        Instruction: This process consists of two steps. The first step is to identify the words in the given sentence that have the highest \\ 
        potential to change the sentiment polarity after substitution, known as the causal words. The second step is to select appropriate \\ 
        replacement words for the causal words that will change the sentiment polarity of the sentence to the desired polarity.\\ Demonstration:\\ Given Sentence: "The movie is the best that I have ever seen."\\ Current Sentiment Polarity: "positive"\\ Target Sentiment Polarity: "negative"\\ Revised Sentence: "The movie is the most wonderful that I have ever seen."\\ Based on the given task definition and instruction, complete the following text by imitating the given demonstration.\\ Given Sentence: "but it also has many of the things that made the first one charming ."\\ Current Sentiment Polarity: "positive"\\ Target Sentiment Polarity: "negative"\\ Revised Sentence:\end{tabular} \\ 
        \bottomrule
    \end{tabular}
}
\caption{Prompts for counterfactual generation for the SA task (with unresonable demonstration).}

\end{table*}

\begin{table*}[]
\resizebox{\linewidth}{!}{
    \begin{tabular}{l}
        \toprule
        \begin{tabular}[c]{@{}l@{}}
            Task Definition: Revise the hypothesis sentence, using minimal changes, to alter the relationship between it and the premise \\
            sentence to either entailment, contradiction, or neutral.\\ 
            Instruction: This process consists of two steps. The first step is to identify the words in the given hypothesis sentence that \\
            have the highest potential to change the relationship with the premise sentence after substitution, known as the causal words. \\
            The second step is to select appropriate replacement words for the causal words that will change the relationship with the premise \\
            sentence to the desired relationship, either entailment, contradiction, or neutral.\\ 
            Demonstration:\\ Given Premise Sentence: "A group of men riding bicycles in a line."\\ 
            Given Hypothesis Sentence: "The men riding together."\\ Current Relationship between the premise sentence and the hypothesis sentence: "Entailment"\\ Target Relationships: {[}"Contradiction","Neutral"{]}\\ 
            Generated Hypothesis Sentences: {[}\{"target\_relationship":,"Contradiction","revised\_sentence":"The men riding horses."\},\\ 
            \{"target\_relationship":,"Neutral","revised\_sentence":"The men are professionals."\}{]}\\ 
            Based on the given task definition and instruction, complete the following text by imitating the given demonstration.\\ 
            Given Premise Sentence: "A group of young girls in a fenced in area."\\ Given Hypothesis Sentence: "a group of sisters playing"\\ Current Relationship between the premise sentence and the hypothesis sentence: "Neutral"\\ Target Relationships: {[}"Entailment","Contradiction"{]}\\ Generated Hypothesis Sentences:\end{tabular} \\ 
        \bottomrule
    \end{tabular}
}
\caption{Prompts for counterfactual generation for the NLI task.}
\end{table*}

\begin{table*}[]
\resizebox{\linewidth}{!}{
    \begin{tabular}{l}
        \toprule
        \begin{tabular}[c]{@{}l@{}}
            Task Definition: Revise the hypothesis sentence, using minimal changes, to alter the relationship between it and the \\ 
            premise sentence to either entailment, contradiction, or neutral.\\ Instruction: This process consists of two steps. \\ 
            The first step is to identify the words in the given hypothesis sentence that have the highest potential to change the \\ 
            relationship with the premise sentence after substitution, known as the causal words. The second step is to select \\ 
            appropriate replacement words for the causal words that will change the relationship with the premise sentence to the \\ 
            desired relationship, either entailment, contradiction, or neutral.\\ Demonstration:\\ Given Premise Sentence: "A group of men riding bicycles in a line."\\ Given Hypothesis Sentence: "The men riding together."\\ Current Relationship between the premise sentence and the hypothesis sentence: "Entailment"\\ 
            Causal Words Identification: The relationship "Entailment" depends on words "riding together" in the hypothesis sentence.\\ Target Relationship: "Contradiction"\\ 
            Causal Words Replacement: To change the relationship between the premise sentence and the hypothesis sentence from \\
            "Entailment" to "Contradiction", causal words "riding together" are replaced by "riding horses".\\ Revised Sentence: "The men riding horses."\\ 
            Target Relationship: "Neutral"\\ 
            Causal Words Replacement: To change the relationship between the premise sentence and the hypothesis sentence from \\ 
            "Entailment" to "Neutral", causal words "riding together" are replaced by "are professionals".\\ 
            Revised Sentence: "The men are professionals."\\ 
            Based on the given task definition and instruction, complete the following text by imitating the given demonstration.\\ 
            Given Premise Sentence: "A group of young girls in a fenced in area."\\ 
            Given Hypothesis Sentence: "a group of sisters playing"\\ 
            Current Relationship between the premise sentence and the hypothesis sentence: "Neutral"\\ 
            Causal Words Identification: \\ 
            Target Relationship: "Entailment"\\ Causal Words Replacement: \\ Revised Sentence: \\ Target Relationship: "Contradiction"\\ Causal Words Replacement: \\ Revised Sentence:\end{tabular} \\ 
            \bottomrule
    \end{tabular}
}
\caption{Prompts for counterfactual generation for the NLI task (with chain-of-thought).}
\end{table*}

\begin{table*}[]
\resizebox{\linewidth}{!}{
    \begin{tabular}{l}
        \toprule
        \begin{tabular}[c]{@{}l@{}}
            Task Definition: Revise the hypothesis sentence, using minimal changes, to alter the relationship between it and the \\ 
            premise sentence to either entailment, contradiction, or neutral.\\ Instruction: This process consists of two steps. \\ 
            The first step is to identify the words in the given hypothesis sentence that have the highest potential to change the \\ 
            relationship with the premise sentence after substitution, known as the causal words. The second step is to select \\ 
            appropriate replacement words for the causal words that will change the relationship with the premise sentence to the \\ 
            desired relationship, either entailment, contradiction, or neutral.\\ Demonstration:\\ Given Premise Sentence: "A group of men riding bicycles in a line."\\ Given Hypothesis Sentence: "The men riding together."\\ Current Relationship between the premise sentence and the hypothesis sentence: "Entailment"\\ Target Relationships: {[}"Contradiction","Neutral"{]}\\ 
            Generated Hypothesis Sentences: {[}\{"target\_relationship":,"Contradiction","revised\_sentence":"The men riding bicycles."\},\\ 
            \{"target\_relationship":,"Neutral","revised\_sentence":"The men are riding horses."\}{]}\\ 
            Based on the given task definition and instruction, complete the following text by imitating the given demonstration.\\ 
            Given Premise Sentence: "A group of young girls in a fenced in area."\\ Given Hypothesis Sentence: "a group of sisters playing"\\ 
            Current Relationship between the premise sentence and the hypothesis sentence: "Neutral"\\ 
            Target Relationships: {[}"Entailment","Contradiction"{]}\\ 
            Generated Hypothesis Sentences:\end{tabular} \\ 
            \bottomrule
    \end{tabular}
}
\caption{Prompts for counterfactual generation for the NLI task (with unresonable demonstration).}
\end{table*}

\begin{table*}[]
\resizebox{\linewidth}{!}{
    \begin{tabular}{l}
        \toprule
        \begin{tabular}[c]{@{}l@{}}
            Task definition: Generate words that can replace entities in the given sentence, whose type is the same as the original \\
            entity type, and refer to the demonstration for the output format.\\ 
            Demonstration:\\ 
            Given Sentence: "Apple was founded in 1978." \\ 
            Given Entities: {[}\{"entity\_span":"Apple","entity\_type":"organization"\},\{"entity\_span":"1978","entity\_type":"date"\}{]} \\ 
            Replaceable Entity Words: {[}\{"entity\_span":"Apple","entity\_type":"organization","replaceable\_entity\_words":{[}"Google","OpenAI","Microsoft"{]}\}, \\ 
            \{"entity\_span":"1978","entity\_type":"date"\},"replaceable\_entity\_words":{[}"1978","1890","March"{]}{]}\\ 
            Based on the given task definition and instruction, complete the following text by imitating the given demonstration.\\ 
            Given Sentence: "Cargill thinks that even though the merchant has a contract stating that it wo n't bring this cocoa to \\
            market until after March 1991 , there is some evidence the contract has been modified ."\\ 
            Given Entities: {[}\{"entity\_span":"Cargill","entity\_type":"PERSON"\}, \{"entity\_span":"March 1991","entity\_type":"DATE"\}{]}\\ Replaceable Entity Words:\end{tabular} \\ 
            \bottomrule
    \end{tabular}
}
\caption{Prompts for counterfactual generation for the NER task.}
\end{table*}

\begin{table*}[]
\resizebox{\linewidth}{!}{
    \begin{tabular}{l}
        \toprule
        \begin{tabular}[c]{@{}l@{}}
            Task Definition: Generate words that can replace entities in the given sentence, whose type is the same as the original \\
            entity type, and refer to the demonstration for the output format.\\ 
            Demonstration: \\ 
            Given Sentence: "Apple was founded in 1978." \\ 
            Given Entities: {[}\{"entity\_span":"Apple","entity\_type":"organization"\},\{"entity\_span":"1978","entity\_type":"date"\}{]} \\ 
            Replaceable Entity Words: {[}\{"entity\_span":"Apple","entity\_type":"organization","replaceable\_entity\_words":{[}"1970","1878","March"{]}\}, \\ 
            \{"entity\_span":"1978","entity\_type":"date","replaceable\_entity\_words":{[}"Google","Microsoft","OpenAI"{]}\}{]}\\ 
            Based on the given task definition and instruction, complete the following text by imitating the given demonstration.\\ Given Sentence: "- Prime minister names former general Avraham Tamir to staff after failing to establish national security council ."\\ Given Entities: {[}\{"entity\_span":"Avraham Tamir","entity\_type":"PER"\}{]}\\ Replaceable Entity Words:\end{tabular} \\ 
            \bottomrule
            
    \end{tabular}
}
\caption{Prompts for counterfactual generation for the NER task (with unresonable demonstration).}
\end{table*}

\begin{table*}[]
\resizebox{\linewidth}{!}{
    \begin{tabular}{l}
        \toprule
        \begin{tabular}[c]{@{}l@{}}
            Task Definition: Revise a given sentence with minimal changes to change the relation between the head and tail entity.\\ 
            Instruction: This process involves three steps. The first step is to identify context words (excluding entity words) in \\ 
            the given sentence that are most likely to change the relation between the head and tail entity when replaced, known as \\ 
            the causal words. The second step is to select a potential target relation from the candidate relation set, which must \\ 
            conform the relevant commonsense of head and tail entity. The third step is to replace the causal words with appropriate \\ 
            words to change the original relation into potential target relations.\\ 
            Note: The found potential target relation must belong to the candidate relation set. If there are no potential target  \\ 
            relation that conforms the commonsense, just output None.\\ 
            Demonstration: Given Sentence: "the key is moved into a chest."\\ Head entity: "key"\\ Tail entity: "chest"\\ 
            Relation between the Head and Tail entity: "entity-destination"\\ 
            Candidate Relation Set: \{"message-topic", "topic-message", "destination-entity", "content-container", "container-content", \\ 
            "effect-cause", "cause-effect", "whole-component", "component-whole", "collection-member", "member-collection", \\ 
            "agency-instrument", "instrument-agency", "producer-product", "product-producer", "entity-origin", "origin-entity"\}\\ 
            Revised Sentence: \{"target\_relation":"entity-origin","revised\_sentence":"the key is from a chest."\}\\ 
            Based on the given task definition and instruction, complete the following text by imitating the given demonstration.\\ 
            Given Sentence: "The series comprises some re-issues of the previous books , as well as new titles ."\\ 
            Head entity: "titles"\\ Tail entity: "series"\\ Relation between the Head and Tail entity: "Component-Whole"\\ 
            Candidate Relation Set: \{"Instrument-Agency", "Member-Collection", "Cause-Effect", "Entity-Destination", "Content-Container", \\
            "Message-Topic", "Product-Producer", "Entity-Origin", "Whole-Component", "Agency-Instrument", "Collection-Member", \\
            "Effect-Cause", "Destination-Entity", "Container-Content", "Topic-Message", "Producer-Product", "Origin-Entity", "Other"\}\\ 
            Revised Sentence:\end{tabular} \\ 
            \bottomrule
    \end{tabular}
}
\caption{Prompts for counterfactual generation for the RE task.}
\end{table*}

\begin{table*}[]
\resizebox{\linewidth}{!}{
    \begin{tabular}{l}
        \toprule
        \begin{tabular}[c]{@{}l@{}}
            Task Definition: Revise a given sentence with minimal changes to change the relation between the head and tail entity.\\ 
            Instruction: This process involves three steps. The first step is to identify context words (excluding entity words) in \\ 
            the given sentence that are most likely to change the relation between the head and tail entity when replaced, known as \\ 
            the causal words. The second step is to select a potential target relation from the candidate relation set, which must \\ 
            conform the relevant commonsense of head and tail entity. The third step is to replace the causal words with appropriate \\ 
            words to change the original relation into potential target relations.\\ 
            Note: The found potential target relation must belong to the candidate relation set. If there are no potential target \\ 
            relation that conforms the commonsense, just output None.\\ 
            Demonstration: Given Sentence: "the key is moved into a chest."\\ 
            Head entity: "key"\\ Tail entity: "chest"\\ Relation between the Head and Tail entity: "entity-destination"\\ 
            Candidate Relation Set: \{message-topic, topic-message, destination-entity, content-container, container-content,\\  
            effect-cause, cause-effect, whole-component, component-whole, collection-member, member-collection, \\ 
            agency-instrument, instrument-agency, producer-product, product-producer, entity-origin, origin-entity\}\\ 
            Causal Words Identification: The relation type "entity-destination" depends on contextual words "moved into".\\ 
            Potential Relation Discovery: The relation between "key" and "chest" can be "entity-origin".\\ 
            Causal Words Replacement: To change the relation between "key" and "chest" from "entity-destination" to "entity-origin", \\
            causal words "moved into" can be replaced by "from".\\ 
            Revised Sentence: \{"target\_relation":"entity-origin","revised\_sentence":"the key is from a chest."\}\\ 
            Based on the given task definition and instruction, complete the following text by imitating the given demonstration.\\ 
            Given Sentence: "Tom Thabane , who set up the All Basotho Convention four months ago , said his party would do \\ 
            more against the poverty that wracks the southern African nation ."\\ Head entity: "All Basotho Convention"\\ Tail entity: "Tom Thabane"\\ Relation between the Head and Tail entity: "org:founded\_by"\\ 
            Candidate Relation Set: \{"Instrument-Agency", "Member-Collection", "Cause-Effect", "Entity-Destination", "Content-Container", \\
            "Message-Topic", "Product-Producer", "Entity-Origin", "Whole-Component", "Agency-Instrument", "Collection-Member", \\
            "Effect-Cause", "Destination-Entity", "Container-Content", "Topic-Message", "Producer-Product", "Origin-Entity", "Other"\}\\ 
            Causal Words Identification: \\ Potential Relation Discovery: \\ Causal Words Replacement: \\ Revised Sentence:\end{tabular} \\ 
            \bottomrule
    \end{tabular}
}
\caption{Prompts for counterfactual generation for the RE task (with chain-of-thought).}
\end{table*}

\begin{table*}[]
\resizebox{\linewidth}{!}{
    \begin{tabular}{l}
        \toprule
        \begin{tabular}[c]{@{}l@{}}
        Task Definition: Revise a given sentence with minimal changes to change the relation between the head and tail entity.\\ 
        Instruction: This process involves three steps. The first step is to identify context words (excluding entity words) in \\ 
        the given sentence that are most likely to change the relation between the head and tail entity when replaced, known as \\ 
        the causal words. The second step is to select a potential target relation from the candidate relation set, which must \\ 
        conform the relevant commonsense of head and tail entity. The third step is to replace the causal words with appropriate \\ 
        words to change the original relation into potential target relations.\\ 
        Note: The found potential target relation must belong to the candidate relation set. If there are no potential target \\
        relation that conforms the commonsense, just output None.\\ Demonstration: Given Sentence: "the key is moved into a chest."\\ 
        Head entity: "key"\\ Tail entity: "chest"\\ Relation between the Head and Tail entity: "entity-destination"\\ 
        Candidate Relation Set: \{"message-topic", "topic-message", "destination-entity", "content-container", "container-content", \\ 
        "effect-cause", "cause-effect", "whole-component", "component-whole", "collection-member", "member-collection", \\ 
        "agency-instrument", "instrument-agency", "producer-product", "product-producer", "entity-origin", "origin-entity"\}\\ 
        Revised Sentence: \{"target\_relation":"topic-message","revised\_sentence":"the key causes a chest."\}\\ 
        Based on the given task definition and instruction, complete the following text by imitating the given demonstration.\\ 
        Given Sentence: "Fine workmanship is the result almost entirely of the worker 's accurate eye and deft hand ."\\ Head entity: "eye"\\ Tail entity: "worker"\\ 
        Relation between the Head and Tail entity: "Component-Whole"\\ 
        Candidate Relation Set: \{"Instrument-Agency", "Member-Collection", "Cause-Effect", "Entity-Destination", "Content-Container", \\
        "Message-Topic", "Product-Producer", "Entity-Origin", "Whole-Component", "Agency-Instrument", "Collection-Member", \\
        "Effect-Cause", "Destination-Entity", "Container-Content", "Topic-Message", "Producer-Product", "Origin-Entity", "Other"\}\\ 
        Revised Sentence:\end{tabular} \\ 
        \bottomrule
    \end{tabular}
}
\caption{Prompts for counterfactual generation for the RE task (with unresonable demonstration).}
\label{prompt_re_3}
\end{table*}

\begin{table}[]
    \centering
    \small
    \resizebox{0.5\textwidth}{!}{
    \begin{tabular}{ll}
        \toprule
        \multicolumn{2}{l}{{CoNLL2003}} \\
        \midrule
        PER & person \\
        LOC & location \\
        ORG & organization \\
        MISC & miscellaneous \\
        \midrule
        \multicolumn{2}{l}{{OntoNotes}} \\
        \midrule
        CARDINAL&cardinal\\
        DATE&date\\
        EVENT&event\\
        FAC&facility\\
        GPE&country city state\\
        LANGUAGE&language\\
        LAW&law\\
        LOC&location\\
        MONEY&monetary\\
        NORP&nationality religious political group\\
        ORDINAL&ordinal\\
        ORG&organization\\
        PERCENT&percent\\
        PERSON&person\\
        PRODUCT&product\\
        QUANTITY&quantity\\
        TIME&time\\
        WORK\_OF\_ART&work of art\\
        \midrule
        \multicolumn{2}{l}{{SemEval}} \\
        \midrule
        Entity-Destination & destined for \\
        Cause-Effect & lead to \\
        Content-Container & contained in \\
        Other & other \\
        Entity-Origin & originated from \\
        Member-Collection & belongs to \\
        Product-Producer & made by \\
        Component-Whole & composed of \\
        Message-Topic & pertaining to \\
        Instrument-Agency & performed by \\
        \midrule
        \multicolumn{2}{l}{{TACRED}} \\
        \midrule
        per:age & age \\
        org:founded & founded\\
        per:date\_of\_birth & date of birth\\
        per:country\_of\_birth & country of birth\\
        org:alternate\_names & alternate names\\
        org:founded\_by & founded by\\
        per:cause\_of\_death & cause of death\\
        org:country\_of\_headquarters & country of headquarters\\
        per:alternate\_names & alternate names\\
        org:members & members\\
        per:cities\_of\_residence & cities of residence\\
        org:city\_of\_headquarters & city of headquarters\\
        org:political/religious\_affiliation & political religious affiliation\\
        per:employee\_of & employee of\\
        per:stateorprovinces\_of\_residence & state or provinces of residence\\
        org:member\_of & member of\\
        org:stateorprovince\_of\_headquarters & state or province of headquarters\\
        per:parents & parents\\
        org:dissolved & dissolved\\
        org:parents & parents\\
        per:children & children\\
        per:spouse & spouse\\
        per:date\_of\_death & date of death\\
        per:city\_of\_death & city of death\\
        per:countries\_of\_residence & countries of residence\\
        org:top\_members/employees & top members employees\\
        no\_relation & no relation\\  
        per:title & title\\
        per:schools\_attended & schools attended\\
        per:religion & religion\\
        per:siblings & siblings\\
        per:charges & charges\\
        per:origin & origin\\
        per:other\_family & other family\\
        per:stateorprovince\_of\_death & state or province of death\\
        org:website & website\\
        per:stateorprovince\_of\_birth & state or province of birth\\
        org:shareholders & shareholders\\
        org:subsidiaries & subsidiaries\\
        per:city\_of\_birth & city of birth\\
        org:number\_of\_employees/members & number of employees members\\
        per:country\_of\_death & country of death \\
        \bottomrule
    \end{tabular}
    }
    \caption{The label-word mapping of NER and RE tasks for BART-based SLMs.}
    \label{tab:label-word}
\end{table}

\end{document}